\documentclass{article} 
\usepackage[table]{xcolor}
\usepackage[preprint]{colm2025_conference}
\usepackage{multirow}
\usepackage{microtype}
\usepackage{url}
\usepackage{booktabs}
\usepackage{makecell}
\usepackage{graphicx}
\usepackage[font=small]{caption} 
\usepackage{subfig}
\usepackage{amssymb}
\usepackage{pifont}
\usepackage[most]{tcolorbox}
\usepackage{wrapfig}
\usepackage{float}
\usepackage{longtable}
\usepackage{setspace}
\usepackage{listings}
\usepackage{verbatim}
\usepackage{footnote}
\usepackage{CJKutf8}
\usepackage{kotex}
\usepackage{spverbatim}

\usepackage{hyperref}

\usepackage{tikz}
\usetikzlibrary{tikzmark,positioning}

\definecolor{darkblue}{rgb}{0, 0, 0.5}
\definecolor{yellow}{RGB}{255, 230, 100}
\hypersetup{colorlinks=true, citecolor=darkblue, linkcolor=darkblue, urlcolor=darkblue}

\newcommand{\cmark}{\ding{51}}%
\newcommand{\xmark}{\ding{55}}%

\title{Kanana: Compute-efficient Bilingual Language Models}

\author{\textbf{Kanana LLM Team} \thanks{A detailed contributor list can be found in the last section of the main paper.} \\
\texttt{kanana-llm@kakaocorp.com} \\
}

\begin{document}

\ifcolmsubmission
\linenumbers
\fi

\maketitle

\begin{abstract}
We introduce Kanana, a series of bilingual language models that demonstrate exceeding performance in Korean and competitive performance in English. 
The computational cost of Kanana is significantly lower than that of state-of-the-art models of similar size.
The report details the techniques employed during pre-training to achieve compute-efficient yet competitive models, including high quality data filtering, staged pre-training, depth up-scaling, and pruning and distillation.
Furthermore, the report outlines the methodologies utilized during the post-training of the Kanana models, encompassing supervised fine-tuning and preference optimization, aimed at enhancing their capability for seamless interaction with users.
Lastly, the report elaborates on plausible approaches used for language model adaptation to specific scenarios, such as embedding, retrieval augmented generation, and function calling.
The Kanana model series spans from 2.1B to 32.5B parameters with 2.1B models (base, instruct, embedding) publicly released to promote research on Korean language models.
\end{abstract}

\section{Introduction}
\begin{figure}[ht]
    \centering
    \includegraphics[width=0.95\textwidth]{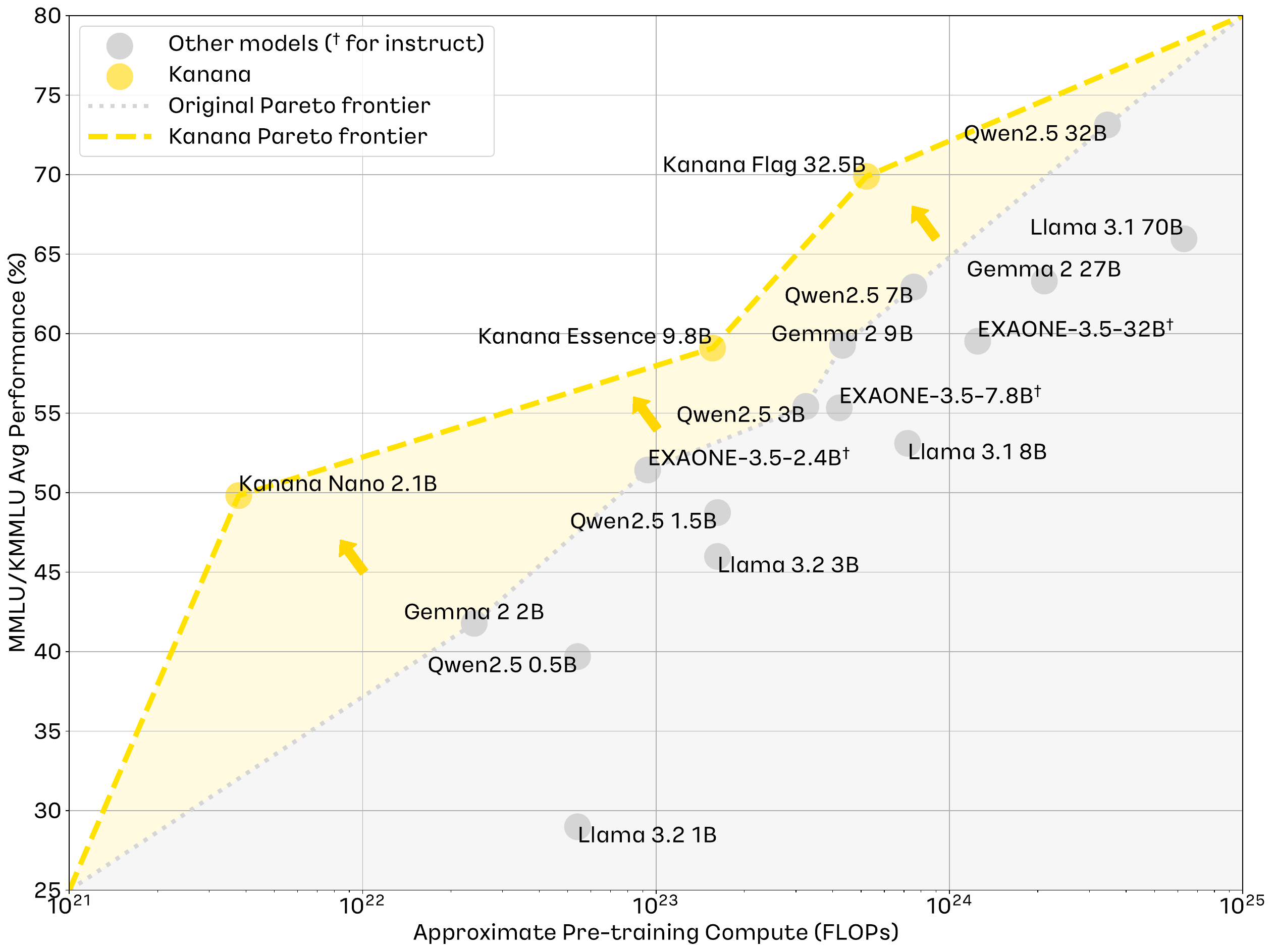}
    \caption{
    Performance to pre-training computational cost for Kanana and comparable models.
    We measure computational cost in FLOPs (Floating Point Operations), which is approximately calculated as 6 $\times$ training tokens $\times$ model size \citep{kaplan2020scalinglaws}.
    We only calculate student training FLOPs for distillation models.
    Obviously, Kanana models achieves decent performance given their limited computational cost.
    }
    \label{fig:flops-vs-mmlu}
\end{figure}

Recent breakthroughs in large language models (LLMs) have been driven by increasing training data~\citep{chinchilla} and model parameters~\citep{gpt3, kaplan2020scalinglaws, palm}.
However, advances have also introduced substantial computational costs that reach millions of dollars \citep{llama3}, which poses a challenge to the community on developing LLMs \textit{from scratch}.
As a result, reducing computational cost has emerged as a crucial problem in order to popularize the development of LLMs for both academia and industry \citep{zhao2024galore, fishman2025scaling-fp8, wang2025fp4-training}. 
To this end, recent works have presented various solutions to the computation problem in model architectures and scaling \citep{deepseek-v2, kim2023solar, muralidharan2024compact}, through data \citep{fineweb-edu, ask-llm}, and through training strategies \citep{deepseek-llm, minicpm}.

As the product of our endeavor to address the computational challenges, we introduce \textit{Kanana} model family, developed using only a fraction of computational cost while maintaining performance compared to those of the state-of-the-art (SOTA) open LLMs.
The family of models includes pre-trained base model and post-trained instruction models in sizes of \{2.1B, 9.8B, 32.5B\}. 
We show in \autoref{fig:flops-vs-mmlu} that Kanana models establish a new Pareto frontier in the computational cost of the train time versus the performance.

In the pre-training phase, as it accounts for the majority of the training costs for LLMs, we focus on reducing its computational demands while maintaining performance.
Since the cost of the pre-training phase primarily arises from the large dataset size and model scale, we reduce it by improving both data efficiency and training efficiency.
To improve data efficiency, we carefully curate a training dataset of 3 trillion tokens, enabling our models to achieve competitive performance despite using a smaller dataset than SOTA pre-trained models.
For training efficiency, we employ cost-effective techniques such as staged pre-training \citep{minicpm, ibrahim2024simple-and-scalable} and depth up-scaling \citep{kim2023solar} to reduce computational costs associated with model size.
From the models obtained, we extend pruning and distillation technique \citep{muralidharan2024compact} to train smaller models using only a handful subset of the pre-training data.

Leveraging the strong performances of Kanana base models, we further develop instruction and domain-specific adaptation models.
To develop instruction models, we apply a post-training process that includes supervised fine-tuning and preference optimization.
As a result, our instruction models achieve competitive performance to that of SOTA models on various tasks, including English/Korean chat, general knowledge reasoning, instruction following, code generation, and mathematical problem-solving.
In addition, we adapt instruction models to develop embedding models, retrieval-augmented generation models, and function-calling models.

\section{Pre-training}

Since pre-training constitutes the majority of computational costs, we focus on reducing the expenses of this stage and show our results in Section \ref{subsec:pretrain_overview}.
To enhance efficiency in pre-training LLMs, we employ two key strategies: data efficiency and training efficiency.
In Section \ref{subsec:pretrain_data}, we discuss our data curation method to maximize the data efficiency under fixed token budget.
In Section \ref{subsec:pretrain_train_process}, we adopt cost-effective training techniques to minimize the computational overhead associated with model scaling.

\subsection{Performance}
\label{subsec:pretrain_overview}

\begin{table}[h]
\centering
\resizebox{\columnwidth}{!}{
\begin{tabular}{l|cccccc|c}
\toprule
    \multirow{2}{*}{\textbf{Models}} & \textbf{MMLU} & \textbf{KMMLU} & \textbf{HAE-RAE} & \textbf{HumanEval} & \textbf{MBPP} & \textbf{GSM8K} & \multirow{2}{*}{\textbf{Avg}} \\
    & \textit{5-shot} & \textit{5-shot} & \textit{5-shot} & \textit{0-shot} & \textit{3-shot} & \textit{5-shot} & \\
\midrule
\rowcolor{yellow} Kanana Flag 32.5B & 77.68 & 62.10 & \textbf{90.47} & \textbf{51.22} & 63.40 & 70.05 & 69.15 \\
Qwen2.5 32B & \textbf{83.10} & \textbf{63.15} & 75.16 & 50.00	& \textbf{73.40} & \textbf{82.41} & 71.20 \\
Gemma 2 27B & 75.45 & 51.16 & 69.11 & \textbf{51.22} & 64.60 & 74.37 & 64.32 \\
EXAONE-3.5-32B$^\dag$ & 72.68 & 46.36 & 82.22 & - & - & - & - \\
Aya Expanse 32B$^\dag$ & 74.52 & 49.57 & 80.66 & - & - & - & - \\
\midrule
\rowcolor{yellow} Kanana Essence 9.8B & 67.61 & 50.57 & \textbf{84.97} & 40.24 & 53.60 & 63.61 & 60.10 \\
Llama 3.1 8B & 65.18 & 41.02 & 61.78 & 35.37 & 48.60 & 50.87 & 50.47 \\
Qwen2.5 7B & \textbf{74.19} & \textbf{51.68} & 67.46 & \textbf{56.71} & \textbf{63.20} & \textbf{83.85} & 66.18 \\
Gemma 2 9B & 70.34 & 48.18 & 66.18 & 37.20 & 53.60 & 68.16 & 57.28 \\
EXAONE-3.5-7.8B$^\dag$ & 65.36 & 45.30 & 77.54 & - & - & - & - \\
Aya Expanse 8B$^\dag$ & 62.52 & 40.11 & 71.95 & - & - & - & - \\
\midrule
\rowcolor{yellow} Kanana Nano 2.1B & 54.83 & 44.80 & \textbf{77.09} & 31.10 & 46.20 & 46.32 & 50.06 \\
Llama 3.2 3B & 56.40 & 35.57 & 47.66 & 25.61 & 39.00 & 27.37 & 38.60 \\
Qwen2.5 3B & \textbf{65.57} & \textbf{45.28} & 61.32 & \textbf{37.80} & \textbf{55.60} & \textbf{69.07} & 55.77 \\
Gemma 2 2B & 52.89 & 30.67 & 45.55 & 20.12 & 28.20 & 24.72 & 33.69 \\
EXAONE-3.5-2.4B$^\dag$ & 59.27 & 43.58 & 68.65 & - & - & - & - \\
\midrule\midrule
Llama 3.1 70B & 78.93 & 53.00 &	76.35 & 57.32 &	66.60 &	81.73 & 68.99 \\
Qwen2.5 72B & 86.12 & 68.57 & 80.84 & 55.49 & 76.40 & 92.04 & 76.58 \\ 
\bottomrule
\end{tabular}
}
\caption{
Performance of Kanana base models on a set of standard benchmarks. 
The best scores are denoted in bold. 
70B sized Models have been included for reference purposes.
$^\dag$ For these models, results are obtained using instruct models because base model checkpoints are not released. 
}\label{table:base-eval-1}
\end{table}

We evaluate our pre-trained models using a series of standard benchmarks designed to assess English/Korean general knowledge, code, and mathematical reasoning.
For general knowledge, we employ multiple choice tasks of MMLU \citep{hendryckstest2021} for English knowledge, and KMMLU \citep{son2024kmmlu} and HAE-RAE \citep{son-etal-2024-hae} for Korean-specific knowledge.
To evaluate domain-specific abilities, we use HumanEval \citep{humaneval} and MBPP \citep{mbpp} for code and GSM8K \citep{cobbe2021gsm8k} for mathematical reasoning.
We use log-likelihood for multiple choice tasks, and greedy generation for generative tasks.

To demonstrate the effectiveness of our training strategy, we compare our models with representative open-source models in various model sizes \citep{llama3, gemma2024gemma2, qwen25techreport, research2024exaone, dang2024ayaexpansecombiningresearch}.
For EXAONE and Aya Expanse models \citep{research2024exaone, dang2024ayaexpansecombiningresearch}, we report only the performances on multiple-choice tasks using the same evaluation protocol.
This decision is based on the observation that multiple-choice performances largely remain unchanged between the base and instruct models, whereas generative tasks exhibit notable divergences (see Appendix \ref{appendix:pre-vs-post} for a detailed discussion).

As shown in \autoref{table:base-eval-1} and \autoref{fig:flops-vs-mmlu}, our models demonstrate strong performance in various domains and exhibit impressive Korean language capabilities, while requiring significantly less training compute. 
Kanana Flag 32.5B outperforms Llama 3.1 70B, Gemma 2 27B, and EXAONE-3.5-32B on knowledge-intensive natural language understanding benchmarks, such as MMLU and KMMLU, while consuming substantially fewer computational resources. 
In particular, the computational cost is even lower than that of Llama 3.1 8B, and is similar to Gemma 2 9B and EXAONE-3.5-7.8B. 
On the HAE-RAE benchmark, all Kanana LLMs demonstrate superior performance compared to other LLMs of similar sizes.
\subsection{Data}
\label{subsec:pretrain_data}

We train Kanana models on 3 trillion tokens, primarily focusing on English and Korean bilingual capabilities.
We collect our corpora from various sources and categorize them as English web, Korean web, academic, code, encyclopedic documents, and instruction data. 
All our data come from publicly available sources and do not include data from Kakao's products or services.

We begin by collecting various open-source datasets from multiple high-quality sources such as arXiv and Wikipedia.
However, we observe that these datasets often suffer quality issues due to suboptimal extraction pipelines, resulting in omissions or incoherent paragraph ordering (see Appendix \ref{appendix:suboptimal-arxiv-wiki} for details).
Inherently, we improve source-specific extraction processes for these sources and re-extract documents with more valuable information and higher coherence.
For code datasets, we utilize open-source datasets from \citet{starcoder} and \citet{starcoder2}. We use only permissively licensed code and exclude any with non-permissive or missing licenses.
Following \citet{inf-llm}'s observation that adding instruction data at the end of pre-training enhances performance after SFT, we also incorporate instruction data with decontamination.

Utilizing the high potential of web as a source of valuable and diverse documents \citep{dclm, nemotron-cc, deepseekmath}, we apply series of filtering methods to extract high quality data.
The first filtering process is cascaded filtering pipeline \citep{yi, llama3, dclm, gemma, fineweb-edu} consisting of deduplication, heuristic filtering, and personally identifiable information (PII) anonymization. 
After the cascaded filtering, we further apply language-specific model-based filtering on high quality documents \citep{nemotron-cc, deepseekmath, dclm, fineweb-edu} separately on English and Korean. 
For English web documents, we utilize a DCLM \citep{dclm} classifier.
For Korean web documents, due to the lack of publicly available high quality classifiers, we iteratively train edu filter as high quality classifier using \texttt{FastText} \citep{fasttext} based on the FineWeb-Edu pipeline \citep{fineweb-edu}.
When applying the FineWeb-Edu pipeline, we observe that most of the documents are classified as uneducational, leading to a distribution imbalance.
To address this issue, we iteratively retrain the classifier by augmenting educational documents from the previous iteration.

To assess the quality of our edu filter and Korean web corpus, we perform experiments by continual pre-training Llama 3 8B with 25B tokens.
As shown in \autoref{tab:kor-edu}, the quality of our Korean web corpus is comparable to that of FineWeb 2 \citep{penedo2024fineweb-2}, which is the largest open-source Korean corpus.
Furthermore, when using our edu filter to extract high quality data from Korean web corpus, we observe a significant performance improvement in the experimental results through training.
Interestingly, we observe that using high quality English data, regardless of the quality of Korean data, can improve the scores on Korean benchmarks such as KMMLU and HAE-RAE, as well as the English benchmark MMLU. 
The results from this experiment make a foundation of our intuition for data mixture strategy in the staged pre-training in the following section.

\begin{table}[ht]
    \centering
        \begin{tabular}{ll|ccc}
        \toprule
        \multirow{2}{*}{\textbf{English Corpus}} & \multirow{2}{*}{\textbf{Korean Corpus}} & \textbf{MMLU} & \textbf{KMMLU} & \textbf{HAE-RAE} \\
        & & \textit{5-shot} & \textit{5-shot} & \textit{5-shot} \\
        \midrule
        - & - & 65.14 & 40.29 & 61.23 \\ 
        \midrule
        DCLM random & FineWeb2 Korean & 64.16 & 41.02 & 70.39 \\
        DCLM random & Our Korean web & 63.59 & 41.41 & 71.31 \\
        DCLM random & Our Korean web w/ edu filter & 63.47 & 43.60 & 74.89\\
        DCLM high & FineWeb2 Korean & 65.36 & 41.78 & 71.22 \\
        DCLM high  & Our Korean web & 64.80 & 41.96 & 72.59 \\
        DCLM high & Our Korean web w/ edu filter & \textbf{65.40} & \textbf{44.19} & \textbf{75.99} \\
        
        \bottomrule
        \end{tabular}
    \caption{Performance of Llama 3 8B before and after continual pre-training with only 25B tokens, using different combinations of English and Korean corpora at a 1:1 ratio.}
    \label{tab:kor-edu}
\end{table}

In summary, we share two insights to consider when building bilingual corpora with underrepresented language for enhanced computational efficiency.
(1) Prioritize quality over quantity.
For languages that do not have vast tokens available, such as Korean, prioritizing quality over quantity is an effective solution.
(2) Knowledge from English data transfers to Korean.
Even with quality filtering on Korean dataset, English data remains a primary source of diverse and high-quality knowledge.
We observe that, under the same conditions for the quality of Korean data, improving the quality of English data leads to higher scores on Korean-related benchmarks.
\subsection{Training Process}
\label{subsec:pretrain_train_process}

To enhance computational efficiency in pre-training LLMs, we employ three key techniques: staged pre-training from scratch, depth up-scaling, and pruning and distillation.
In Section \ref{subsec:pretrain_staged_pretrain}, we first train 26.8B and 8B models using a staged pre-training approach, which serves as the foundation for obtaining LLMs at various scales.
In Section \ref{subsec:pretrain_dus}, we describe the process to obtaining \textit{Kanana Flag 32.5B} and \textit{Kanana Essence 9.8B} models by depth up-scaling from 26.8B and 8B models, respectively.
In Section \ref{subsec:pretrain_pd}, we derive \textit{Kanana Nano 2.1B} model through pruning and distillation from the 8B model, reducing training costs while achieving superior performance compared to training a model from scratch.

\subsubsection{Staged Pre-training from Scratch}
\label{subsec:pretrain_staged_pretrain}

\begin{figure}[h]
    \centering
    \subfloat[Stage 1 data]{
        \includegraphics[width=0.45\textwidth]{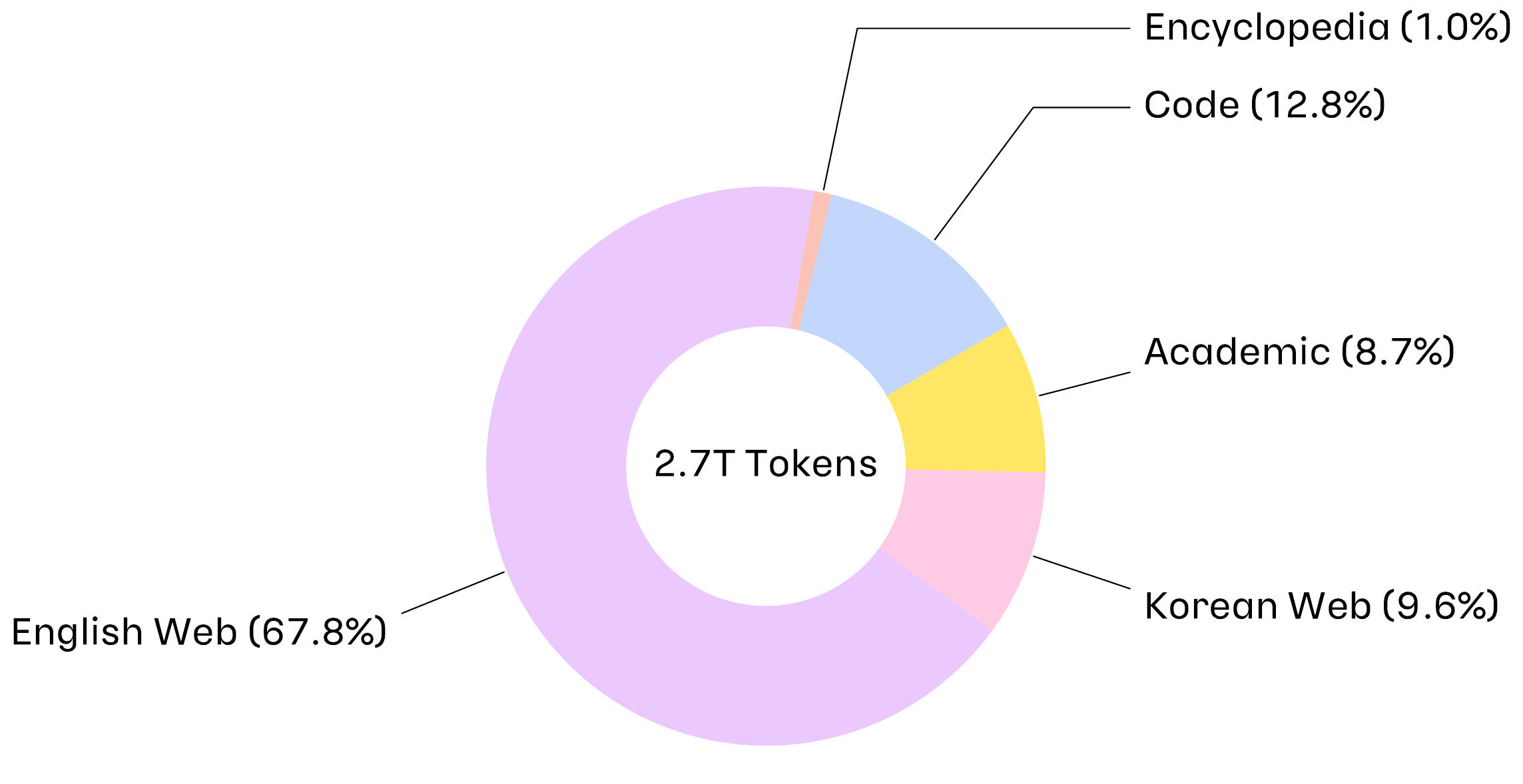}
        \label{fig:pre_stage1_data}
    }
    \hfill
    \subfloat[Stage 2 data]{
        \includegraphics[width=0.45\textwidth]{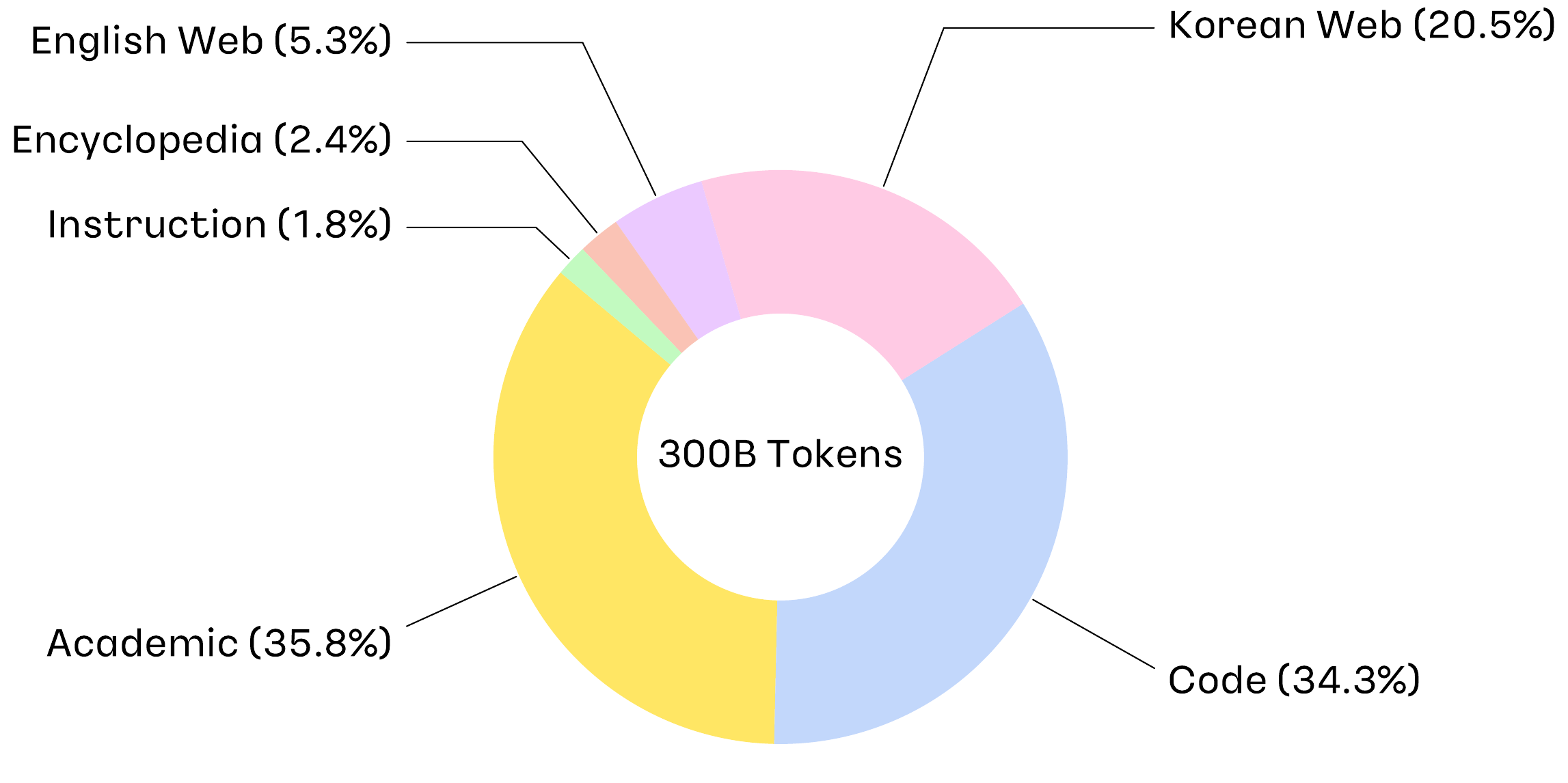}
        \label{fig:pre_stage2_data}
    }
    \caption{Kanana's staged pre-training data mixture.}
    \label{fig:pre_data_stats}
\end{figure}

\begin{table}[h]
\centering
\resizebox{\columnwidth}{!}{
\begin{tabular}{lc|cccccc|c}
\toprule
    \multirow{2}{*}{Models} & \multirow{2}{*}{Stage} & \textbf{MMLU} & \textbf{KMMLU} & \textbf{HAE-RAE} & \textbf{HumanEval} & \textbf{MBPP} & \textbf{GSM8K} & \multirow{2}{*}{\textbf{Avg}} \\
    & & \textit{5-shot} & \textit{5-shot} & \textit{5-shot} & \textit{0-shot} & \textit{3-shot} & \textit{5-shot} & \\
\midrule
\multirow{2}{*}{26.8B} & Stage 1 & 73.38 & 54.26 & 84.97 & 32.32 & 47.20 & 57.77 & 58.32 \\
& Stage 2 & 74.27 & 59.04 & 88.45 & 51.22 & 61.60 & 67.48 & 67.01 \\
\midrule
\multirow{2}{*}{8B} & Stage 1 & 63.48 & 45.51 & 77.27 & 23.78 & 35.80 & 35.03 & 46.81 \\
& Stage 2 & 64.22 & 48.30 & 83.41 & 40.24 & 51.40 & 57.09 & 57.44 \\
\bottomrule
\end{tabular}
}
\caption{
Performance of from-scratch Kanana models at the end of each training stage.
} \label{table:from-scratch}
\end{table}

To maximize performance under fixed compute budget, we adopt the staged pre-training strategy \citep{minicpm, gemma, opencoder, yi-lightning, granite} with two stages.
Staged pre-training divides the pre-training process into multiple stages, starting with training LLMs on a large amount of moderate-quality data in the initial stages, and gradually increasing the proportion of high quality data in the subsequent stages. 

We begin by training 8B from scratch using the diverse 2.7 trillion in stage 1 as shown in Figure \autoref{fig:pre_stage1_data}.
In stage 2, we further train the model using 300 billion tokens shown in Figure \ref{fig:pre_stage2_data}. 
Specifically, we set aside high quality data for each category using the available high quality classifiers.
Then, we perform lightweight annealing experiments to select candidate datasets to search for the data mixture following \citet{llama3}.
Then, the optimal data mixture is selected through ablation study.
The final model of stage 2 results in a 2.79 point increase in KMMLU and a 10.63 point increase in average performance, demonstrating the effectiveness and efficiency of staged pre-training. 
We apply the same data mixture that was used during the training of 8B to 26.8B model. 
Direct application of the recipe consistently yields remarkable performance and stable training as shown in \autoref{table:from-scratch}, demonstrating the scalability of our recipe. See Appendix \ref{appendix:pretrain-details} for our pre-training configurations.

\subsubsection{Depth Up-scaling}
\label{subsec:pretrain_dus}

To further enhance the model performance within limited resources after pre-training, we adopt the depth up-scaling (DUS) which increases model capacity by stacking additional layers \citep{kim2023solar}.
We apply DUS to expand Kanana 8B into Kanana Essence 9.8B and Kanana 26.8B into Kanana Flag 32.5B.
After the up-scaling process, each model variant is further trained on the same data mixtures used in pre-training, with 100 billion tokens dedicated to stage 1 and another 100 billion to stage 2.
Results of the up-scaling strategy demonstrates that the additional layers consistently contribute to performance enhancements as summarized in \autoref{tab:performance}.
\begin{table}[ht]
    \centering
    \resizebox{\columnwidth}{!}{%
    \begin{tabular}{l|cccccc|c}
    \toprule
    \multirow{2}{*}{\textbf{Models}} & \textbf{MMLU} & \textbf{KMMLU} & \textbf{HAE-RAE} & \textbf{HumanEval} & \textbf{MBPP} & \textbf{GSM8K} & \multirow{2}{*}{\textbf{Avg}} \\
    & \textit{5-shot} & \textit{5-shot} & \textit{5-shot} & \textit{0-shot} & \textit{3-shot} & \textit{5-shot} & \\
    \midrule
    26.8B + DUS (32.5B) & 77.68 & 62.10 & 90.47 & 51.22 & 63.40 & 70.05 & 69.15 \\
    26.8B & 74.27 & 59.04 & 88.45 & 51.22 & 61.60 & 67.48 & 67.01 \\
    \midrule
    8B + DUS (9.8B)     & 67.61 & 50.67 & 84.98 & 40.24 & 53.60 & 63.61 & 60.10 \\
    8B & 64.22 & 48.30 & 83.41 & 40.24 & 51.40 & 57.09 & 57.44 \\
    \bottomrule
    \end{tabular}%
    }
    \caption{Performance comparison of Kanana models before and after depth up-scaling.}
    \label{tab:performance}
\end{table}

\autoref{tab:performance} illustrates the performance improvements achieved through depth up-scaling.
Kanana Essence 9.8B consistently outperforms its non-upscaled version, Kanana 8B with the average score rising from 57.52 to 60.12. 
This improvement is evident in MMLU, KMMLU, HAE-RAE, MBPP, and GSM8K, except for HumanEval. 
Similarly, Kanana Flag 32.5B achieves average score of 69.15, notably surpassing the non-upscaled Kanana 26.8B model. 
These results emphasize the effectiveness of depth up-scaling in improving various benchmark scores.

Notably, our strategy saves 11.06\% of total computational cost compared to the training of 9.8B and 32.5B LLMs from scratch.
This strategy of increasing model capacity through depth up-scaling only occupies about 6.67\% of the total computing resources across the entire training procedure. 
In combination with pre-training, depth up-scaling offers a strategic approach to significantly enhance model performance without introducing heavy computational demands of building new models from scratch.

\subsubsection{Pruning and Distillation}
\label{subsec:pretrain_pd}

In opposition to efficiently up-scaling the model size, knowledge distillation is an effective method to efficiently down-scale the model size \citep{hinton2015knowledge-distillation, gunter2024apple, llama3.2}.
Leveraging the 8B model from Section \ref{subsec:pretrain_staged_pretrain}, we efficiently produce smaller models by improving the pruning and distillation of Minitron \citep{muralidharan2024compact, sreenivas2024llm}.
This process allows us to produce models with better performance at one-tenth of the data size compared to training from scratch, as shown in \autoref{tab:pd-vs-fs}.
We further show that iteratively extending the process beyond two iterations remains effective, preserving 87-99\% of KMMLU score at only 50\% of the model size, as shown in \autoref{tab:pd-iterative}.
Our models achieve competitive performance to recent open-source models \citep{allal2025smollm2smolgoesbig, llama3, gemma2024gemma2, qwen25techreport}, as presented in \autoref{tab:pd-base-all}.

\begin{table}[ht]
    \centering
    \resizebox{\columnwidth}{!}{%
    \begin{tabular}{l|c|cccccc|c}
    \toprule
    \multirow{2}{*}{\textbf{Models}} & \multirow{2}{*}{\textbf{\makecell{Training \\ Tokens}}} & \textbf{MMLU} & \textbf{KMMLU} & \textbf{HAERAE} & \textbf{HumanEval} & \textbf{MBPP} & \textbf{GSM8K} & \multirow{2}{*}{\textbf{Avg}} \\
    & & \textit{5-shot} & \textit{5-shot} & \textit{5-shot} & \textit{0-shot} & \textit{3-shot} & \textit{5-shot} & \\
    \midrule
    2.1B PD & 0.3T & 54.83 & 44.80 & 77.09 & 31.10 & 46.20 & 46.32 & 50.06 \\
    2.1B & 3T & 50.66 & 36.61 & 68.74 & 24.45 & 41.60 & 36.69 & 43.13 \\
    \bottomrule
    \end{tabular}%
    }
    \caption{
    Token consumption and performance of pruning \& distillation (PD) from preceding models and training from scratch. We use the same 2.1B architecture.
    }
    \label{tab:pd-vs-fs}
\end{table}

\begin{table}[ht]
    \centering
    \begin{tabular}{l|cccccc|c}
    \toprule
    \multirow{2}{*}{\textbf{Models}} & \textbf{MMLU} & \textbf{KMMLU} & \textbf{HAERAE} & \textbf{HumanEval} & \textbf{MBPP} & \textbf{GSM8K} & \multirow{2}{*}{\textbf{Avg}} \\
    & \textit{5-shot} & \textit{5-shot} & \textit{5-shot} & \textit{0-shot} & \textit{3-shot} & \textit{5-shot} \\
    \midrule
    \tikzmark{t} 8B$^\dag$ & 64.22 & 48.30 & 83.41 & 40.24 & 51.40 & 57.09 & 57.44 \\
    \tikzmark{a} 4.5B & 59.74 & 48.09 & 82.58 & 34.76 & 48.60 & 57.01 & 55.13 \\
    \tikzmark{b} 2.1B & 54.83 & 44.80 & 77.09 & 31.10 & 46.20 & 46.32 & 50.06 \\
    \tikzmark{c} 1.3B & 53.55 & 39.91 & 72.59 & 28.05 & 39.60 & 36.01 & 44.95 \\
    \tikzmark{d} 635M & 46.28 & 34.60 & 62.69 & 23.17 & 31.40 & 19.26 & 36.23 \\
    \tikzmark{e} 385M & 41.16 & 31.70 & 47.94 & 18.90 & 24.00 & 10.83 & 29.08 \\
    \tikzmark{f} 192M & 26.11 & 30.16 & 19.71 & 12.80 & 12.40 & 2.43  & 17.27 \\
    \bottomrule
    \end{tabular}%
    \caption{Performance through iterative compression beyond two iterations. Each model is pruned from the preceding model. $^\dag$ Each model is distilled using the 8B model as the teacher.}
    \label{tab:pd-iterative}
    \begin{tikzpicture}[overlay, remember picture, shorten >=1pt, shorten <=1pt, transform canvas={yshift=.25\baselineskip}]
        \draw [-stealth] ({pic cs:t}) [bend right=50] to ({pic cs:a});
        \draw [-stealth] ({pic cs:a}) [bend right=50] to ({pic cs:b});
        \draw [-stealth] ({pic cs:b}) [bend right=50] to ({pic cs:c});
        \draw [-stealth] ({pic cs:c}) [bend right=50] to ({pic cs:d});
        \draw [-stealth] ({pic cs:d}) [bend right=50] to ({pic cs:e});
        \draw [-stealth] ({pic cs:e}) [bend right=50] to ({pic cs:f});
    \end{tikzpicture}
\end{table}

In order to improve the pruning and distillation process, we refine Minitron's width importance scoring while preserving its simplicity and efficiency.
Its scoring process begins by measuring the importance of embedding channels, feed-forward neurons, and attention heads using activations from a small calibration dataset.
Next, we show that summing layer-wise scores plays a crucial role in performance, whereas the prior work performed ablations along batch and sequence axes.
Moreover, for Grouped-Query Attention (GQA) \citep{ainslie2023gqa}, we improve performance by ensuring query-key-value alignment. Specifically, we remove an equal number of query heads within each group, as shown in \autoref{fig:pd-gqa}.
Additionally, since Kanana employs SwiGLU \citep{swiglu}, we choose between averaging gate and up states or using intermediate states, whereas the original formulation relies on pre-activation values.
All ablation results for importance scoring are in \autoref{tab:pd-score-detail}.

We further enhance the pruning strategies with a focus on intermediate model structures.
Consistent with the findings from Minitron, we observe that excessive single-step compression leads to significant degradation.
Although maintaining attention heads is generally beneficial, our experiments reveal that pruning them for smaller models is effective when done earlier at larger scales as presented in \autoref{tab:pd-attn-heads}.
Additionally, we find that input and output embeddings can be tied by averaging without causing noticeable degradation, which we apply when pruning from 4.5B to 2.1B as shown in \autoref{tab:pd-tie}.

Lastly, we observe that the composition of distillation data directly influences the performance, while pruning data is less important.
For models larger than 2B, we use high-quality 300 billion tokens of stage 2 described in Section \ref{subsec:pretrain_staged_pretrain}.
However, for smaller models, increasing the proportion of general-domain English data increases both English performance and other benchmark scores, as shown in \autoref{tab:pd-distill-data}.

In conclusion, our comprehensive pre-training process, which includes staged pre-training, depth up-scaling, and iterative pruning and distillation, offers a compute-efficient strategy for developing high-performing language models. 
This combined approach not only enhances performance across diverse benchmarks, but also ensures computational efficiency, demonstrating the effectiveness of our strategy in producing a robust family of models spanning the range from 2.1B to 32.5B. See Appendix \ref{appendix:pd-details} for our pruning and distillation configurations.

\section{Post-training}

Building on Kanana pre-trained models, we further develop instruction-tuned models for direct interaction by natural language.
In Section \ref{subsec:post_train_performance}, we highlight the performance of Kanana instruction-tuned models, demonstrating superior performance on Korean tasks and competitive results on other tasks.
Section \ref{subsec:post-training-data} presents the details of the specifics regarding the Supervised Fine-Tuning (SFT) and preference datasets.
Section \ref{subsec:post_train_process} outlines the extensive post-training techniques applied on Kanana instruction models.

\subsection{Performance}
\label{subsec:post_train_performance}

\begin{table}[h]
\centering
\resizebox{0.95\columnwidth}{!}{
\begin{tabular}{l|cccc|c}
\toprule
\multirow{2}{4em}{\textbf{Models}} & \multicolumn{4}{c|}{\textit{Chat}} & \textit{Instruction Following} \\
      & \textbf{MT-Bench} & \textbf{LogicKor} & \textbf{KoMT-Bench} & \textbf{WildBench} & \textbf{IFEval} \\
\midrule
\rowcolor{yellow} Kanana Flag 32.5B & 8.356 & \textbf{9.524} & \textbf{8.058} & 54.14 & \textbf{0.856} \\
Qwen2.5 32B & 8.331 & 8.988 & 7.847 & 51.13 & 0.822 \\
Gemma 2 27B & 8.088 & 8.869 & 7.373 & 46.46 & 0.817 \\
EXAONE-3.5-32B & \textbf{8.375} & 9.202 & 7.907 & \textbf{54.30} & 0.845 \\
Aya Expanse 32B & 7.788 & 8.941 & 7.626 & 48.36 & 0.735 \\
\midrule
\rowcolor{yellow} Kanana Essence 9.8B & 7.769 & 8.964 & 7.706 & 47.27 & 0.799 \\
Llama 3.1 8B & 7.500 & 6.512 & 5.336 & 33.20 & 0.772 \\
Qwen2.5 7B & 7.625 & 7.952 & 6.808 & 41.31 & 0.760 \\
Gemma 2 9B & 7.633 & 8.643 & 7.029 & 40.92 & 0.750 \\
EXAONE-3.5-7.8B & \textbf{8.213} & \textbf{9.357} & \textbf{8.013} & \textbf{50.98} & \textbf{0.826} \\
Aya Expanse 8B & 7.131 & 8.357 & 7.006 & 38.50 & 0.645\\
\midrule
\rowcolor{yellow} Kanana Nano 2.1B &  6.400 & 7.964 & 5.857 & 25.41 & 0.720 \\
Llama 3.2 3B & 7.050 & 4.452 & 3.967 & 21.91 & 0.767 \\
Qwen2.5 3B & 6.969 & 6.488 & 5.274 & 25.76 & 0.355 \\
Gemma 2 2B & 7.225 & 5.917 & 4.835 & 28.71 & 0.428 \\
EXAONE-3.5-2.4B & \textbf{7.919} & \textbf{8.941} & \textbf{7.223} & \textbf{41.68} & \textbf{0.790} \\
\midrule\midrule
Llama 3.1 70B & 8.275 & 8.250 & 6.970 & 46.50 & 0.875 \\
Qwen2.5 72B & 8.619 & 9.214 & 8.281 & 55.25 & 0.861 \\
\bottomrule
\end{tabular}
}
\caption{
Performance of Kanana and previous instruction-tuned models in general chat and instruction following benchmarks.
Across all \textit{Chat} benchmarks, we use \texttt{gpt-4o-2024-08-06} as a judge model.
The best scores are denoted in \textbf{bold}.
70B sized models have been included for reference purposes.
}\label{table:chat-eval-2}
\end{table}

\begin{table}[h]
\centering
\resizebox{\columnwidth}{!}{
\begin{tabular}{l|ccc|cc|cc}
\toprule
\multirow{2}{4em}{\textbf{Models}} & \multicolumn{3}{c|}{\textit{General}} & \multicolumn{2}{c|}{\textit{Coding}} & \multicolumn{2}{c}{\textit{Mathematics}} \\
      & \textbf{MMLU} & \textbf{KMMLU} & \textbf{HAE-RAE} &  \textbf{HumanEval+} & \textbf{MBPP+} & \textbf{GSM8K} & \textbf{MATH} \\
\midrule
\rowcolor{yellow} Kanana Flag 32.5B & 81.08 & \textbf{64.19} & \textbf{68.18} & 77.44 & 69.84 & 90.83 & 57.82 \\
Qwen2.5 32B & \textbf{84.40} & 59.37 & 48.30 & \textbf{82.32} & \textbf{71.96} & \textbf{95.30} & \textbf{81.90} \\
Gemma 2 27B & 78.01 & 49.98 & 46.02 & 70.12 & 70.90 & 91.05 & 53.80 \\
EXAONE-3.5-32B & 78.30 & 55.44 & 52.27 & 78.66 & 70.90 & 93.56 & 76.80 \\
Aya Expanse 32B & 74.49 & 42.35 & 51.14 & 64.63 & 65.61 & 75.06 & 42.82 \\
\midrule
\rowcolor{yellow} Kanana Essence 9.8B & 70.64 & 50.76 & \textbf{47.16} & 72.56 & 69.05 & 84.91 & 42.24 \\
Llama 3.1 8B & 71.18 & 39.24 & 40.91 & 60.98 & 57.67 & 82.71 & 49.86 \\
Qwen2.5 7B & \textbf{77.23} & 46.87 & 37.50 & 73.78 & \textbf{70.63} & \textbf{91.58} & \textbf{75.22} \\
Gemma 2 9B & 73.47 & 44.47 & 39.77 & 59.76 & 64.55 & 87.72 & 48.10 \\
EXAONE-3.5-7.8B & 72.62 & \textbf{52.09} & 46.02 & \textbf{79.27} & 66.67 & 89.99 & 73.50 \\
Aya Expanse 8B & 61.23 & 35.78 & 39.20 & 42.68 & 56.88 & 78.85 & 30.80 \\
\midrule
\rowcolor{yellow} Kanana Nano 2.1B & 52.48 & \textbf{38.51} & \textbf{33.52} & 63.41 & 62.43 & 72.32 & 29.26 \\
Llama 3.2 3B & 56.09 & 3.07 & 17.05 & 56.71 & 50.26 & 66.57 & 38.18 \\
Qwen2.5 3B & \textbf{69.18} & 38.33 & 32.39 & 67.68 & \textbf{64.02} & \textbf{84.00} & \textbf{65.72} \\
Gemma 2 2B & 57.69 & 6.99 & 7.95 & 35.37 & 45.24 & 49.81 & 21.68 \\
EXAONE-3.5-2.4B & 63.19 & 14.27 & 14.20 & \textbf{70.73} & 59.79 & 83.78 & 64.04 \\
\midrule\midrule
Llama 3.1 70B & 83.48 & 39.08 & 53.41 & 75.61 & 66.40 & 91.66 & 63.98 \\
Qwen2.5 72B & 87.14 & 65.78 & 60.80 & 81.10 & 75.66 & 95.45 & 82.60 \\
\bottomrule
\end{tabular}
}
\caption{
Performance of Kanana post-trained models on a set of standard benchmarks. 
All benchmarks under General category are measured using 0-shot CoT with respective chat-template of each model.
The best scores are denoted in \textbf{bold}.
70B sized models have been included for reference purposes.
}\label{table:chat-eval-1}
\end{table}

We evaluate our instruction-tuned models across various tasks: chat, instruction following, general knowledge, coding, and mathematics and compare their performance to previous instruction-tuned models.
For general chat ability, we use MT-Bench \citep{zheng2023judging}, LogicKor \citep{park2024logickor}, KoMT-Bench \citep{KoMT-Bench}, and WildBench \citep{lin2024wildbench}.
To test instruction following ability, we use IFEval\footnote{We report the average of Prompt-level strict-accuracy and Instruct-level strict-accuracy.}\citep{zhou2023instructionfollowingevaluationlargelanguage}.
For general knowledge tasks, we use MMLU \citep{hendryckstest2021}, KMMLU \citep{son2024kmmlu}, and HAE-RAE\footnote{We report general knowledge category scores in this section.} \citep{son-etal-2024-hae}, with zero-shot chain-of-thought (CoT) \citep{wei2022chainofthought} setting along with the chat template.
Employing zero-shot CoT with the chat template, rather than multi-shot prompts, allows us to evaluate the inherent capabilities of the instruction model, without residual traces from the pre-trained model.
For coding ability, we use HumanEval+ \citep{evalplus} and MBPP+ \citep{evalplus}.
For Mathematical ability, we use GSM8K \citep{cobbe2021gsm8k} and MATH \citep{hendrycksmath2021}.
See Appendix \ref{subsec:evaluation-prompts-for-post-trained-models} for detailed prompts of benchmarks.

\autoref{table:chat-eval-2} and \autoref{table:chat-eval-1} show that our models excel similar sized models on Korean tasks.
The 32.5B model achieves the highest performance in Korean chat tasks (LogicKor, KoMT-Bench) and Korean knowledge tasks (KMMLU, HAE-RAE). 
The 9.8B and 2.1B models rank second in Korean chat tasks and either best or second-best in Korean knowledge tasks.
Additionally, our models exhibit competitive performance across other tasks except in math.

\subsection{Data} \label{subsec:post-training-data}

We collect 1.2M instruction data instances in English and Korean to address both languages.
To ensure that our post-training data can handle diverse human requests, we define five distinct domains and collect prompts from both public datasets and human contributors.
As a result, our dataset comprises 492K instances for \textit{code}, 260K for \textit{math}, 230K for \textit{instruction following}, 120K for \textit{general chat}, and 96K for \textit{safety}.
The safety dataset includes prompts related to ethics, privacy, toxicity, and bias.

\begin{figure}[h]
    \centering
    \subfloat[SFT data]{
        \includegraphics[width=0.45\textwidth]{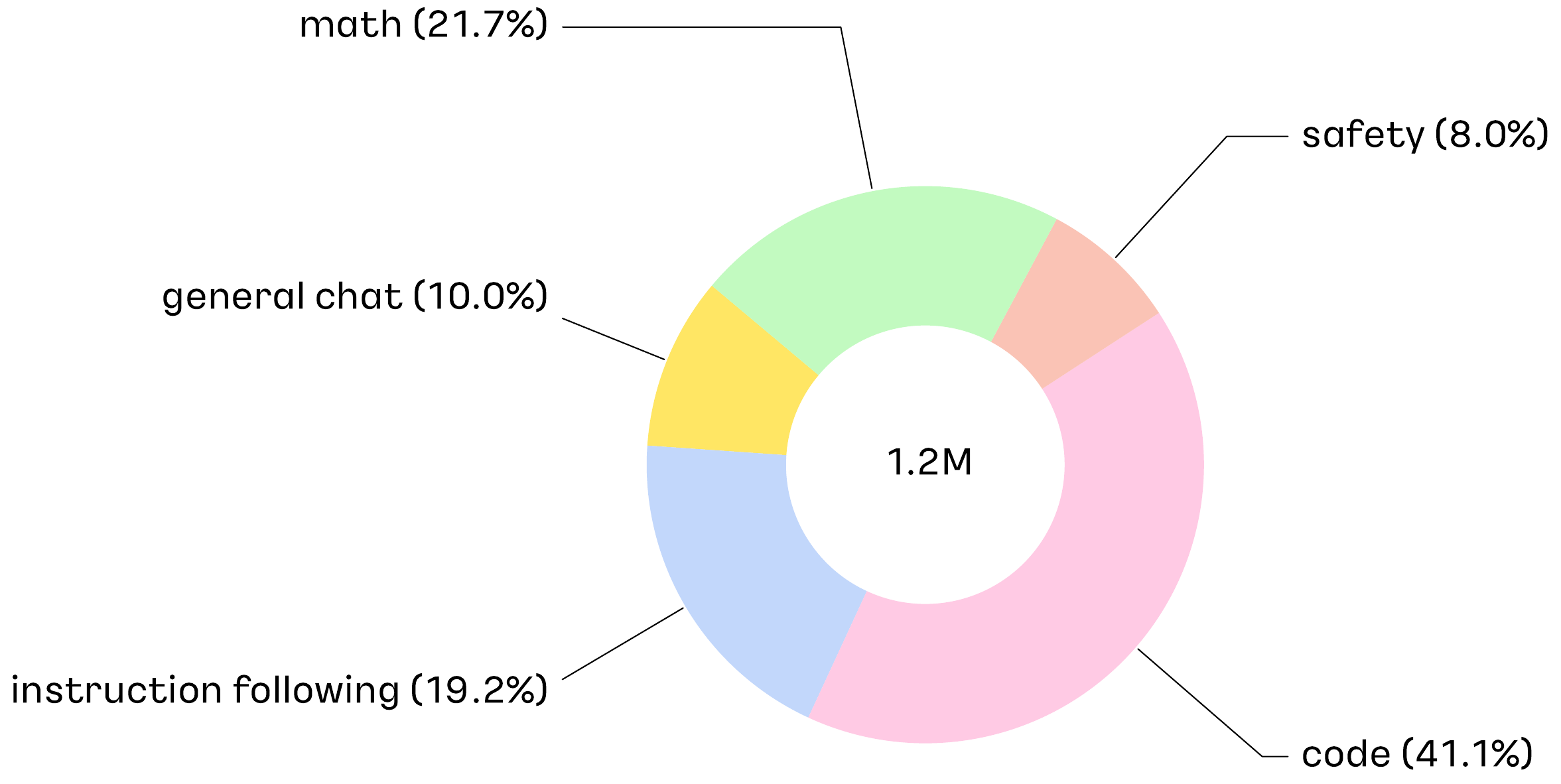}
        \label{fig:post_sft_data}
    }
    \hfill
    \subfloat[Preference data]{
        \includegraphics[width=0.45\textwidth]{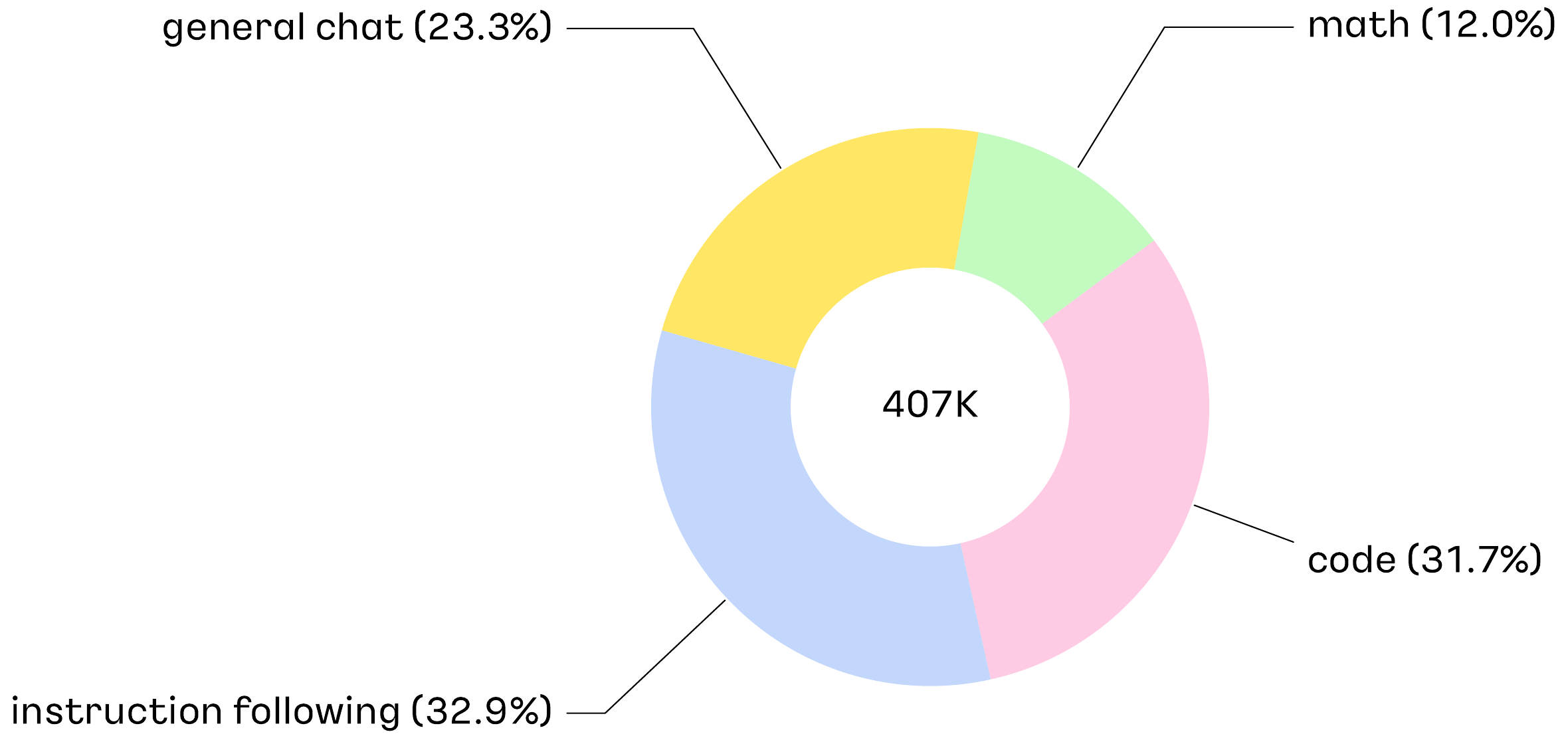}
        \label{fig:post_pref_data}
    }
    \caption{Data size and proportion of each domain.}
    \label{fig:post_data_stats}
\end{figure}

\autoref{fig:post_data_stats} depicts the instance size and proportion of each domain.
For the preference optimization stage, we sub-sampled and balanced the data across each domain.

\subsection{Training Process}
\label{subsec:post_train_process}

We adopt the widely used multi-stage post-training procedure comprising SFT and a series of preference optimization processes \citep{ouyang2022instruct, llama3, qwen25techreport, gemma2024gemma2}.
In Section \ref{subsec:post_train_sft}, we provide details on the SFT process.
In Section \ref{subsec:reward-model-training}, we share information on training our reward model from the SFT model for the subsequent preference optimization process.
In Section \ref{subsec:preference-optimization}, we perform preference optimization on the SFT model, which is a sequential process consisting of offline and online preference optimization.

\begin{figure}[h]
    \centering
    \includegraphics[width=0.8\textwidth]{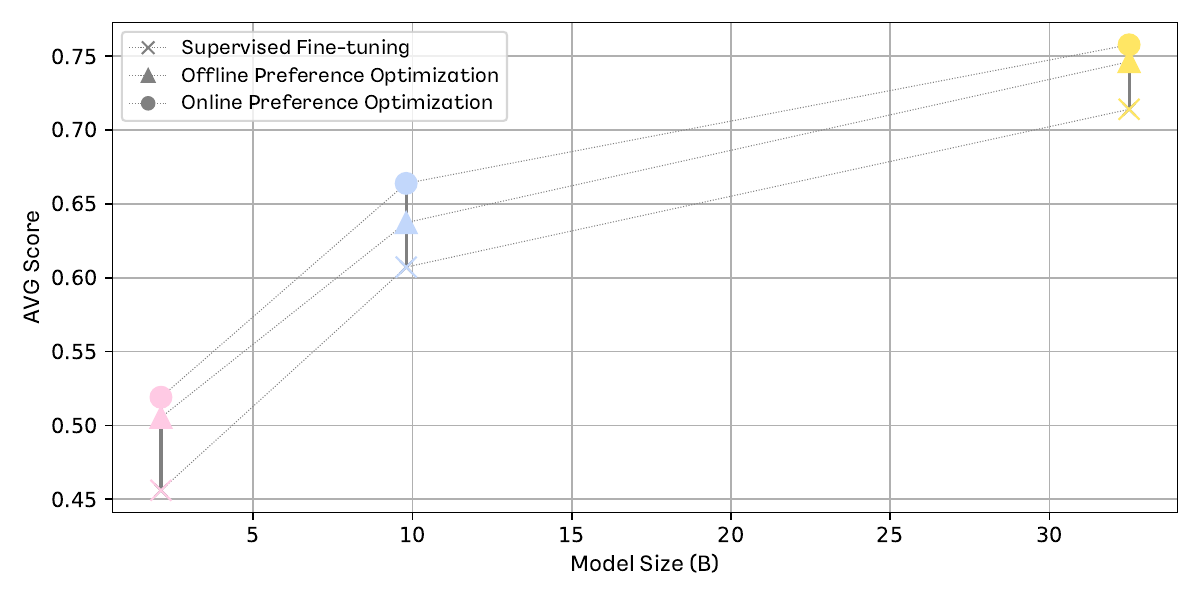}
    \caption{
    Kanana model performance for each stage of training across different model sizes.
    The y-axis is the average of normalized scores of all benchmarks in \autoref{table:chat-eval-2} and \autoref{table:chat-eval-1}.
    The normalization process is done by dividing each score with the maximum possible score.
    }
    \label{fig:stage_score}
\end{figure}

As shown in \autoref{fig:stage_score}, each step of this process quantitatively enhances the instruction-tuned model across different model sizes.
Qualitatively, we observe that during the SFT stage, the model learns to generate structured chat responses while integrating relevant knowledge, and this ability persists through subsequent stages.
Building on the SFT model, the preference optimization stages further enhance performance by refining the model’s tone and manner.
\autoref{sec:qualititive_results} presents qualitative results and illustrates the evolution of model completions throughout each phase of post-training.

\subsubsection{Supervised Fine-Tuning}
\label{subsec:post_train_sft}
\begin{table}[t]
\centering
\resizebox{\columnwidth}{!}{
\begin{tabular}{cccc|cccccc}
\toprule
\multicolumn{4}{c|}{\textbf{Datasets used}} & \multicolumn{5}{c}{\textbf{Normalized Scores}} \\
General & Instruction Following & Code & Math & MT-Bench & IFEval & HumanEval+ & MBPP+ & GSM8K & MATH \\
\midrule
\cmark & \cmark & \cmark & \cmark & 1.00 & 1.00 & 1.00 & 1.00 & 1.00 & 1.00 \\
\midrule
\cmark & \textcolor{red}{\xmark} & \cmark & \cmark & 0.98 & \textcolor{red}{0.72} & 1.06 & 0.99 & 1.03 & 1.07 \\
\cmark & \cmark & \textcolor{red}{\xmark} & \cmark & 0.99 & 1.00 & \textcolor{red}{0.66} & \textcolor{red}{0.72} & 1.01 & 1.05 \\
\cmark & \cmark & \cmark & \textcolor{red}{\xmark} & 0.98 & 1.00 & 1.04 & 1.00 & \textcolor{red}{0.60} & \textcolor{red}{0.59} \\
\bottomrule
\end{tabular}
}
\caption{
Domain mixture ablation for SFT dataset.
All scores are normalized by the score of the SFT model when datasets of all domains have been included in the training set.
We see that removing a specific domain from the training dataset exclusively deteriorates the performance of the respective domain by a significant amount.
}\label{table:sft_abl}
\end{table}

During the SFT stage, the model develops the ability to generate structured chat responses while integrating relevant knowledge.
In this stage, we train the model using 1.2M data instances, as described in Section~\ref{subsec:post-training-data}.
While optimizing the proportion of domain-specific data, we observed that such data is crucial for achieving high performance in its respective domain and does not negatively impact other domains.
\autoref{table:sft_abl} demonstrates that excluding domain-specific data from total dataset only reduces performance on the corresponding domain's benchmark, while performance in other domains remains unaffected.
Consequently, we incorporate the full extent of each domain-specific dataset while ensuring balanced performance across all domains.

\subsubsection{Reward Model Training} \label{subsec:reward-model-training}

We train a reward model for subsequent online preference optimization process, assuming a Bradley-Terry model \citep{bradley1952rank}.
The reward model is trained using the offline preference data along with additional public preference data.
Among various reward models trained with different data proportions and settings, we select the one that demonstrates the strongest best-of-N policy~\citep{gao2023reward-model-overoptimization} performance.
The best-of-N policy performance is evaluated by generating N responses from the policy model, scoring them with the reward model, selecting the highest-scoring response, and then assessing the final response's quality using a benchmark judge.
This approach is based on the intuition that the chosen reward model should effectively evaluate the response distribution of the online preference optimization stage in accordance with the benchmark evaluation criteria.

\subsubsection{Preference Optimization} \label{subsec:preference-optimization}

To further improve the SFT model’s performance on LLM benchmarks, we conduct a preference optimization stage. 
The process begins with offline preference optimization \citep{meng2024simpo, jung2024bco}, where we apply direct preference optimization (DPO)~\citep{rafailov2023direct} using the offline preference data.

We then conduct online preference optimization, initializing from the offline DPO model.
During training, policy-generated responses are evaluated by the reward model from Section \ref{subsec:reward-model-training}, providing training data for online DPO \citep{guo2024online-dpo} with asynchronous response sampling \citep{noukhovitch2024asynchronous-rlhf}.
This approach can be considered as a form of iterative DPO~\citep{xiong2024iterative-preference}.
However, unlike prior work~\citep{tran2023iterative-dpo-snorkel}, we maintain a fixed reference model, specifically the offline DPO model, throughout all iterations.
This decision is based on our observation that updating the reference model led to undesirable increases in response length.

\section{Adaptations}

In this section, we show three examples of practical adaptations of Kanana models to popular applications of LLMs: embedding models, retrieval-augmented models, and function calling models.
Through experimental results, we show that the performances of Kanana models are further improved in each relevant benchmarks when task-specific training techniques are further applied, showcasing the possibility of adapting Kanana models to a wide range of applications.

\subsection{Embedding Models}

Text embeddings, or dense vector representations, are essential for capturing the semantic essence of text \citep{karpukhin2020dense, khattab2020colbert}.
Following the success of LLMs, decoder-only language models have taken their place as a popular backbone of sentence embedding models \citep{muennighoff2022sgpt, wang2023improving, springer2024repetition, ma2024fine, behnamghader2024llm2vec, xu2024bmretriever}.
In this section, we examine the capabilities of the Kanana model, specifically the Kanana Nano 2.1B, as a robust backbone for embedding by employing LLM2Vec \citep{behnamghader2024llm2vec}.
For comparative analysis, we also apply LLM2Vec on models of Llama 3 and Qwen2.5 series with similar model sizes.

\begin{table}[ht]
    \centering
    \resizebox{0.5\columnwidth}{!}{%
    \begin{tabular}{l|ccc}
    \toprule
    \textbf{Embedding Backbone} & {\textbf{English}} & {\textbf{Korean}} & \textbf{Avg} \\
    \midrule
    Kanana Nano 2.1B & 51.56 & \textbf{65.00} & \textbf{58.28} \\
    \midrule
    Llama 3.2 3B & 53.28 & 59.43 & 56.35 \\
    Qwen2.5 3B & \textbf{54.00} & 62.10 & 58.05 \\
    Llama 3.2 1B & 48.77 & 54.68 & 51.73 \\
    Qwen2.5 1.5B & 50.60 & 54.60 & 52.60 \\
    \bottomrule
    \end{tabular}
    }
    \caption{Performance comparison of embedding models on English and Korean retrieval benchmarks. All embedding models are fine-tuned from instruct models. See \autoref{appendix:embedding} for detailed evaluations.}
    \label{tab:embedding-performance}
\end{table}

The embedding models are evaluated on subsets of Massive Text Embedding Benchmark (MTEB) \citep{muennighoff2022mteb} retrieval tasks, including 10 English tasks sourced from the MTEB v2 leaderboard \citep{enevoldsen2025mmtebmassivemultilingualtext} and 8 Korean tasks curated by \citet{KURE}.
\autoref{tab:embedding-performance} presents average nDCG@10 scores for English and Korean, summarizing the performance results on retrieval tasks.

Kanana Nano 2.1B consistently demonstrates competitive performance and serves as an effective backbone for embedding tasks.
As shown in \autoref{tab:embedding-performance}, our 2.1B model not only significantly surpasses Llama 3.2 1B and Qwen2.5 1.5B across both English and Korean benchmarks, but also outperforms Llama 3.2 3B and Qwen2.5 3B on Korean evaluations, despite its smaller size.
Additionally, it achieves a solid English score and the highest average score among the models, highlighting the strong capacity of Kanana Nano 2.1B when fine-tuned for retrieval tasks.

\subsection{Retrieval-Augmented Generation}
Retrieval-Augmented Generation (RAG) methods \citep{lewis2021retrievalaugmentedgenerationknowledgeintensivenlp} enable large language models to access the latest external or proprietary information without altering model parameters \citep{liu2024chatqasurpassinggpt4conversational}.
In order to ensure factual consistency during retrieval, the grounding ability of the model needs to be trained through additional data mixture \citep{lin2024flamefactualityawarealignmentlarge}. 
In this section, we describe a process for developing reliable RAG models with enhanced grounding ability from Kanana LLMs. 

For evaluation, we collect RAG scenario benchmarks and evaluate our model on them.
ContextualBench \citep{nguyen2024sfrragcontextuallyfaithfulllms} is set of multi-hop QA, which we specifically include to consider the conciseness in evaluation. 
FACTs \citep{jacovi2025factsgroundingleaderboardbenchmarking} consists of various tasks with contexts such as reasoning, QA, summarization, rewriting, and extraction. 
\footnote{We filtered with character length of 20k since our base model was trained with token length limit of 8k. This dataset is not labeled golden answer, so we only measure grounding score with it.}
IFEval \citep{zhou2023instructionfollowingevaluationlargelanguage} measures maintenance of helpfulness of our instruct model. 
However, these benchmarks are all English-based, making them insufficient to judge the RAG abilities in Korean. 
To this end, we develop an internal FACTs-like Korean RAG benchmark called RAG-General-Bench that focuses on measuring factual consistency in Korean. 
During the development, human annotators manually constructed the dataset with context, instruction, and reference answer, to evaluate helpfulness as well. 
The benchmark consists of a total of 115 samples with 4 main tasks, categorized into 27 subcategories, providing a diverse set of scenarios for evaluation. There are 2 samples of QA task in \autoref{appendix:rag-bench-example}.


\begin{figure}[h]
    \small
    \centering
    \includegraphics[width=0.6\textwidth]{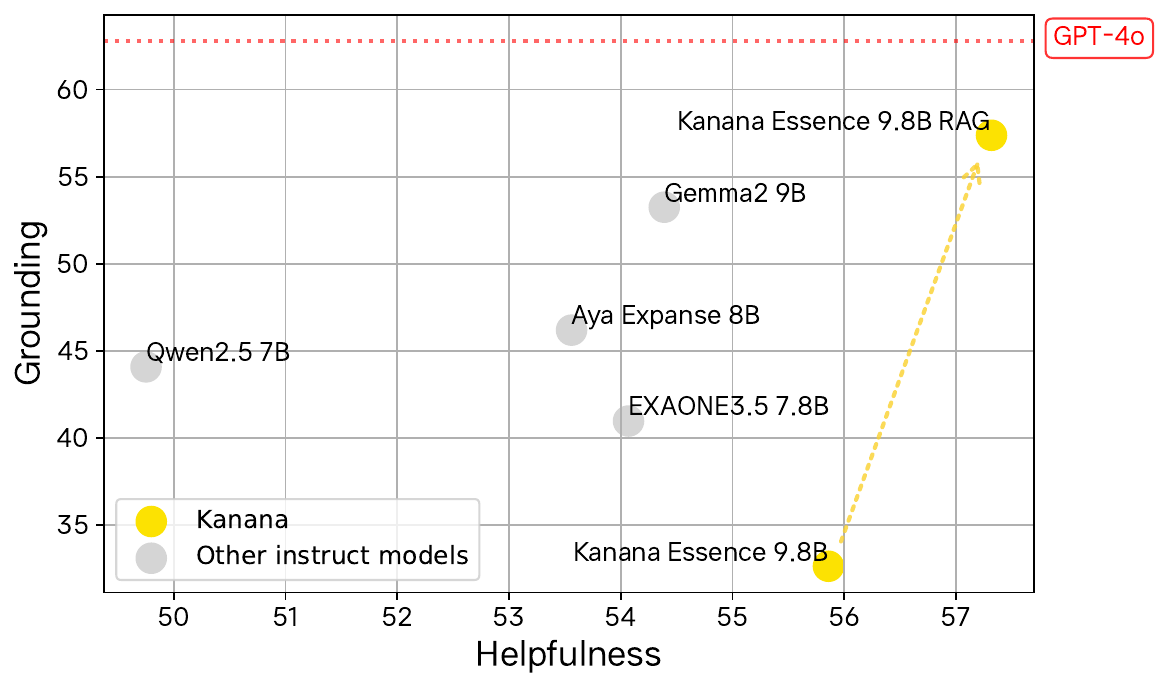}
    \caption{Performance Comparison of Various Models Based on averaged helpfulness and grounding in RAG-General-Bench.}
    \label{fig:rag-main-performance}
\end{figure}

\begin{figure}[h]
\centering
\resizebox{1.0\textwidth}{!}{%
    \includegraphics[width=\textwidth]{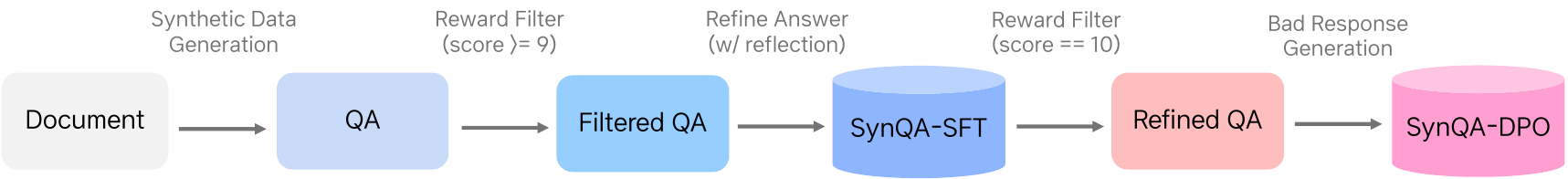}
}
\caption{QA Generation Pipeline}
\label{fig:rag-data-generation-pipeline}
\end{figure}

To increase grounding ability, we synthetically generate question-answer pairs using high-quality bilingual documents as seed documents, following the pipeline in \autoref{fig:rag-data-generation-pipeline}.
Then, we filter out instances with low grounding scores and use LLM-judge to reflect and refine the low grounding instances.
We call the dataset at this point as SynQA-SFT. 
With SynQA-SFT, we augment responses with low grounding score to produce preference dataset that we call SynQA-DPO.
Along with SynQA datasets, we utilize StructLM \citep{zhuang2024structlmbuildinggeneralistmodels} and FollowRAG \citep{dong2024generalinstructionfollowingalignmentretrievalaugmented} to adapt diverse context format and instructions in RAG scenarios and replay SFT dataset from Section \ref{subsec:post-training-data} to prevent general capability of the instruction model from degrading during training.

However, we observe a decline in the helpfulness score as the model is trained through SFT and DPO in \autoref{tab:rag-performance-comparison}.
In order to address this issue, we merge the DPO model with the instruction model to preserve helpfulness \citep{kim2024prometheus2opensource}.
As a result, Kanana Essence 9.8B RAG achieves 91.4\% of GPT-4o's grounding performance while maintaining our instruct model's helpfulness in our benchmark as presented in \autoref{fig:rag-main-performance}.

    

\begin{table}[h]
\centering
\resizebox{\columnwidth}{!}{
    \begin{tabular}{l|c|cc|c|c}
    \toprule
    \textbf{Models} & \multicolumn{1}{c|}{\textbf{FACTs}} & \multicolumn{2}{c|}{\textbf{RAG-General-Bench}} & \multicolumn{1}{c|}{\textbf{ContextualBench}} & \multicolumn{1}{c}{\textbf{IFEval}} \\
    & \small{Grounding} & \small{Grounding} & \small{Helpfulness} & \small{Exact-match} &  \\
    \midrule
    Kanana Essence 9.8B & 40.66 & 32.63 & 55.86 & 20.22 & \textbf{79.93}\\
    \midrule
    + SFT & 62.40& 59.29& 51.60& 48.08& 72.99\\
    + DPO & \textbf{63.09}& \textbf{65.33}& 52.67& \textbf{48.76}& 75.00\\
    + Merge (Kanana Essence 9.8B RAG) & 53.09 & 57.38& \textbf{57.32}& 48.31& 78.44\\
    \bottomrule
    \end{tabular}
}
    \caption{Performance change of each phase of recipe. Grounding score is average of two metric RAGAS \citep{es2023ragasautomatedevaluationretrieval} Faithfulness and rubric based LLM-judge. Helpfulness score is average of two metric RAGAS Answer Relevancy and rubric based LLM-judge. EM means exact matching normalized answer with golden label. IFEval scoring is as same as Section \ref{subsec:post_train_performance}.}
    \label{tab:rag-performance-comparison}
\end{table}

\subsection{Function Calling}
Function calling is an essential ability for large language models (LLMs) to interact with external tools and databases, granting them access to up-to-date information stored in dynamic or structured formats \citep{schick2023toolformer}.
This capability helps 
integrating real-time data with the static knowledge inherent in LLMs, which is particularly vital in enterprises.

Previous works highlight the increasing importance of function calling, which has led to various efforts in data generation for fine-tuning and model evaluation \citep{basu2024api, guo2024api, qin2023toolllm, tang2023toolalpaca, li2023api, rastogi2020towards, liu2024apigen}.
However, these efforts predominantly focused on English, making it necessary to create a function calling dataset for low-resource languages.
To address this gap within Korean contexts, we create a fine-tuning dataset, referred to as \texttt{korean-fc-corpus}.
The corpus is constructed by: (1) translating two key English function calling corpora, glaive-function-calling-v1 (\texttt{gfc-v1}) \citep{glaiveai2023fcv1} and the Schema-Guided Dialogue Dataset (\texttt{sgd}) \citep{rastogi2020towards}, into their Korean equivalents, \texttt{ko-gfc-v1} and \texttt{ko-sgd}; and (2) creating an in-house function calling dataset (\texttt{inhouse-fc}) specifically tailored for corporate applications.

We further adopt two-staged training process comprising domain specific pre-training and supervised fine-tuning to adapt instruct-tuned models to function calling specific tokens and terminologies.
In the domain pre-training phase, we leveraged multiple English-based function calling datasets, including \texttt{gfc-v1}, glaive-function-calling-v2 \citep{glaiveai2024fcv2}, xlam-function-calling-60k \citep{liu2024apigen}, as well as \texttt{sgd}, supplemented by our \texttt{inhouse-fc}.
This foundation enabled us to perform supervised fine-tuning exclusively on \texttt{korean-fc-corpus}.
This two-stage strategy ensures that models become adequately versed in function calling conventions and domain terminologies before focusing on Korean-specific nuances, thereby enhancing their performance in Korean function calling tasks.

\begin{table}[hbt!]
\centering
\resizebox{0.5\columnwidth}{!}{
\begin{tabular}{lcc}
\toprule
\textbf{Models} & \textbf{Single-call} & \textbf{Dialogue} \\
\midrule
Kanana 8B FC & 0.88 & 0.89 \\
gpt-4-0125-preview & 0.94 & 0.94 \\
gpt-4o-2024-05-13 & 0.93 & 0.95 \\
\bottomrule
\end{tabular}
}
\caption{Evaluation on FunctionChat-Bench: Single-call and Dialogue Accuracy}
\label{tab-fc-overall}
\end{table}

To evaluate function calling capabilities in corporate environments, we introduce FunctionChat-Bench \citep{lee2024functionchat}, a benchmark designed for Korean conversational settings.
This benchmark measures performance on two metrics: Single-call accuracy, which evaluates how well a model selects and invokes the necessary function from several options, and Dialogue accuracy, which examines the model’s capability in multi-turn interactions. 
For comparative analysis, we evaluate OpenAI’s proprietary models (
gpt-4-0125-preview, and gpt-4o-2024-05-13) and Kanana 8B FC model as shown in \autoref{tab-fc-overall}.


This result indicates that leveraging task specific fine-tuning on moderately sized LLMs, which are trained at a lower cost, may offer a more cost effective and efficient approach for addressing certain tasks.

\section{Conclusion}

In this report, we present Kanana, a family of large language models available in sizes of \{2.1B, 9.8B, 32.5B\}, with a focus on the cost-effective training procedure compared to other prominent open models.
We emphasize the strong bilingual capability of Kanana models, showcasing state-of-the-art performance on Korean benchmarks of KMMLU, HAE-RAE, and KoMT-Bench and competitive results on various English benchmarks.
However, we also acknowledge the limitations of Kanana models in overall performance on small scale models sizes, particularly in math domains.
To address the limitations, we plan to improve small models and the math ability of all models through data quality and mixture.
To further our commitment in cost-effective training, we intend to explore strategical approaches such as formulating scaling laws and other training methodologies as possible future directions.
Additionally, we aim to expand the linguistic ability from bilingual to multilingual prioritizing the intuition of treating the underrepresented languages covered in this report.
By continuing to build on these efforts, we aspire to make advancements in the field of large language models, balancing performance with efficiency and broadening the linguistic scope of our models.

\clearpage

\let\oldthefootnote\thefootnote
\renewcommand{\thefootnote}{}
\renewcommand{\thefootnote}{\fnsymbol{footnote}}

\section*{Contributors and Acknowledgements}

\subsection*{Pre-training}

Yunju Bak, Doohae Jung, Boseop Kim\footnotemark[2], Nayeon Kim, Hojin Lee, Jaesun Park,  Minho Ryu

\subsection*{Post-training}

Jiyeon Ham, Seungjae Jung, Hyunho Kim, Hyunwoong Ko, Changmin Lee, Daniel Wontae Nam\footnotemark[2], Kyoung-Woon On\footnotemark[2]\footnotemark[3]

\subsection*{Adaptation}

Seulye Baeg, Junrae Cho, Taegyeong Eo, Sunghee Jung, Jieun Kang, EungGyun Kim\footnotemark[2], Eunhwa Kim, Byeongil Ko, Daniel Lee, Donghun Lee, Minchul Lee, Miok Lee, Shinbok Lee, Minho Ryu, Gaeun Seo

\footnotetext[2]{Team leads}
\footnotetext[3]{Work done at Kakao Corp.}

\section*{Acknowledgments}
We thank Donghee Son for the re-extraction of the arXiv data for the pre-training. 
We also thank Gunsoo Han, Jisang Park, and Byeongjae Son for their contributions in the construction of the general chat and code data used in the post-training. 
Finally, we thank Myungchul Shin and Byung-hak Kim for their invaluable support for Kanana models.

\renewcommand{\thefootnote}{\oldthefootnote}

\newpage
\begin{CJK}{UTF8}{mj}
\bibliography{colm2025_conference}
\bibliographystyle{colm2025_conference}
\end{CJK}

\newpage
\appendix
\section{Appendix}
\subsection{Comparison between pre-trained models and post-trained models}\label{appendix:pre-vs-post}

\begin{table}[h]
\centering
\resizebox{\columnwidth}{!}{
\begin{tabular}{lcc|cccccc}
\toprule
    \multirow{2}{*}{\textbf{Models}} & \multirow{2}{*}{\textbf{Tokens}} & \multirow{2}{*}{\textbf{Category}}  & \textbf{MMLU} & \textbf{KMMLU} & \textbf{HAE-RAE} & \textbf{HumanEval} & \textbf{MBPP} & \textbf{GSM8K} \\
    & & & \textit{5-shot} & \textit{5-shot} & \textit{5-shot} & \textit{0-shot} & \textit{3-shot} & \textit{5-shot} \\
\midrule
\midrule
\multirow{2}{*}{Kanana Flag 32.5B}   & \multirow{2}{*}{3.2T} & base     & 77.68 & 62.10 & 90.47 & 51.22 & 63.40 & 70.05 \\
                                     & & instruct & 77.84 & 62.08 & 89.37 & 64.63 & 73.00 & 84.08 \\
\midrule
\multirow{2}{*}{Llama 3.1 70B}       & \multirow{2}{*}{15T} & base     & 78.93 & 53.00 & 76.35 & 57.32 & 66.60 & 81.73 \\
                                     & & instruct & 82.42 & 52.80 & 76.08 & 78.05 & 70.40 & 86.66 \\
\midrule
\multirow{2}{*}{Qwen2.5 32B}         & \multirow{2}{*}{18T} & base     & 83.10 & 63.15 & 75.16 & 50.00 & 73.40 & 82.41 \\
                                     & & instruct & 83.41 & 61.20 & 74.61 & 54.88 & 73.00 & 76.27 \\
\midrule
\multirow{2}{*}{Gemma 2 27B}         & \multirow{2}{*}{13T} & base     & 75.45 & 51.16 & 69.11 & 51.22 & 64.60 & 74.37 \\
                                     & & instruct & 76.39 & 51.49 & 68.84 & 71.34 & 66.20 & 84.46 \\
\midrule
EXAONE-3.5-32B                       & 6.5T & instruct & 72.68 & 46.36 & 82.22 & 74.39 & 67.80 & 55.50  \\
\midrule
Aya Expanse 32B                      & - & instruct & 74.52 & 49.57 & 80.66 & 12.20 & 60.40 & 85.97 \\
\midrule
\midrule
\multirow{2}{*}{Kanana Essence 9.8B} & \multirow{2}{*}{3.2T} & base     & 67.61 & 50.57 & 84.97 & 40.24 & 53.60 & 63.61 \\
                                     & & instruct & 66.45 & 49.95 & 82.95 & 61.59 & 51.60 & 76.04 \\
\midrule
\multirow{2}{*}{Llama 3.1 8B}        & \multirow{2}{*}{15T} & base     & 65.18 & 41.02 & 61.78 & 35.37 & 48.60 & 50.87 \\
                                     & & instruct & 68.17 & 41.22 & 64.44 & 59.76 & 58.00 & 69.52 \\
\midrule
\multirow{2}{*}{Qwen2.5 7B}          & \multirow{2}{*}{18T} & base     & 74.19 & 51.68 & 67.46 & 56.71 & 63.20 & 83.85 \\
                                     & & instruct & 74.23 & 50.13 & 65.72 & 65.85 & 31.60 & 77.56 \\
\midrule
\multirow{2}{*}{Gemma 2 9B}          & \multirow{2}{*}{8T$^\dag$} & base     & 70.34 & 48.18 & 66.18 & 37.20 & 53.60 & 68.16 \\
                                     & & instruct & 72.30 & 46.56 & 66.73 & 56.10 & 57.60 & 80.12 \\
\midrule
EXAONE-3.5-7.8B                      & 9T & instruct & 65.36 & 45.30 & 77.54 & 70.73 & 61.60 & 64.67 \\
\midrule
Aya Expanse 8B                       & - & instruct & 62.52 & 40.11 & 71.95 &  7.93 & 47.40 & 75.97 \\
\midrule
\midrule
\multirow{2}{*}{Kanana Nano 2.1B}    & \multirow{2}{*}{300B$^\dag$} & base     & 54.83 & 44.80 & 77.09 & 31.10 & 46.20 & 46.32 \\
                                     & & instruct & 53.67 & 42.92 & 77.17 & 54.88 & 55.00 & 64.37 \\
\midrule
\multirow{2}{*}{Llama 3.2 3B}        & \multirow{2}{*}{9T$^{\dag\ddag}$} & base     & 56.40 & 35.57 & 47.66 & 25.61 & 39.00 & 27.37 \\
                                     & & instruct & 60.60 & 35.44 & 48.21 & 49.39 & 49.00 & 58.76 \\
\midrule
\multirow{2}{*}{Qwen2.5 3B}          & \multirow{2}{*}{18T} & base     & 65.57 & 45.28 & 61.32 & 37.80 & 55.60 & 69.07 \\
                                     & & instruct & 66.47 & 44.51 & 60.77 & 50.61 & 54.60 & 11.37 \\
\midrule
\multirow{2}{*}{Gemma 2 2B}          & \multirow{2}{*}{2T$^\dag$} & base     & 52.89 & 30.67 & 45.55 & 20.12 & 28.20 & 24.72 \\
                                     & & instruct & 57.04 & 33.48 & 49.77 & 23.78 & 37.80 & 44.05 \\
\midrule
EXAONE-3.5-2.4B                      & 6.5T & instruct & 59.27 & 43.58 & 68.65 & 63.41 & 58.40 & 53.07 \\
\bottomrule
\end{tabular}
}
\caption{
$^\dag$ For distilled models, distillation tokens are only counted
$^\ddag$ Information from \url{https://huggingface.co/meta-llama/Llama-3.2-3B}
}\label{table:appendix-pre-vs-post}
\end{table}

\subsection{Suboptimal extraction of open-source datasets}\label{appendix:suboptimal-arxiv-wiki}

\begin{figure}[h]
\centering
\begin{tcolorbox}[colframe=black!80!white, colback=black!2!white, boxrule=0.5mm, width=\textwidth, arc=2mm, auto outer arc, title=Example of suboptimal extraction from arXiv, fonttitle=\color{white}\bfseries]
\begin{lstlisting}[
  basicstyle=\ttfamily,
  breaklines=true,
  breakatwhitespace=false,
  columns=fullflexible,
  keepspaces=true,
  breakautoindent=false,
  breakindent=0ex
]
(...)
In this work, we use sine and cosine functions of different frequencies:

\begin{align*}
    PE_{(pos,2i)} = sin(pos / 10000^{2i/d_{\text{model}}}) \\
    PE_{(pos,2i+1)} = cos(pos / 10000^{2i/d_{\text{model}}})
\end{align*}

where $pos$ is the position and $i$ is the dimension.  That is, each dimension of the positional encoding corresponds to a sinusoid.  The wavelengths form a geometric progression from $2\pi$ to $10000 \cdot 2\pi$.  We chose this function because we hypothesized it would allow the model to easily learn to attend by relative positions, since for any fixed offset $k$, $PE_{pos+k}$ can be represented as a linear function of $PE_{pos}$.

We also experimented with using learned positional embeddings \citep{JonasFaceNet2017} instead, and found that the two versions produced nearly identical results (see Table~\ref{tab:variations} row (E)).  We chose the sinusoidal version because it may allow the model to extrapolate to sequence lengths longer than the ones encountered during training.

\section{Introduction}

\input{introduction}

\section{Background}

\input{background}

\section{Model Architecture}
\input{model_architecture}

\section{Why Self-Attention}
\input{why_self_attention}

\section{Training}
\input{training}

\section{Results} \label{sec:results}
\input{results}

\section{Conclusion}
In this work, we presented the Transformer, the first sequence transduction model based entirely on attention, replacing the recurrent layers most commonly used in encoder-decoder architectures with multi-headed self-attention.
(...)
\end{lstlisting}
\end{tcolorbox}
\caption{
    Example of suboptimal extraction from arXiv subset of \citet{redpajama}. 
    The original content is from \citet{Vaswani+2017}.
}
\end{figure}

\begin{figure}[h]
\centering
\subfloat[Open-source]{
\begin{tcolorbox}[colframe=black!80!white, colback=black!2!white, boxrule=0.5mm, width=\textwidth, arc=2mm, auto outer arc, title=Example of suboptimal extraction from Wikipedia, fonttitle=\color{white}\bfseries]
\begin{CJK}{UTF8}{mj}
발생원인\\
\\
회전 좌표계\\
\\
회전좌표계
좌표계 x, y, z와 좌표계 x', y', z'을 보자 두 좌표계의 원점은 같다. 각각의 경우에 대해 벡터 .은 두 좌표계에서 다음과 같이 표시된다.\\
\\
. (x, y, z 좌표)\\
\\
. (x', y', z' 좌표계)\\
\\
벡터의 내적을 이용해 x, y, z를 (.), (.), (.)으로 표현할 수 있다. 내적의 방법은 다음과 같다.\\
\\
.\\
\\
.\\
\\
.
\\
으로 표현되는 것을 확인할 수 있다.
\end{CJK}
\end{tcolorbox}
}

\subfloat[Improved]{
\begin{tcolorbox}[colframe=black!80!white, colback=black!2!white, boxrule=0.5mm, width=\textwidth, arc=2mm, auto outer arc, title=Example of improved extraction from Wikipedia, fonttitle=\color{white}\bfseries]
\begin{CJK}{UTF8}{mj}
\#\# 발생원인\\
\\
\#\#\# 회전 좌표계\\
\\
\#\#\#\# 회전좌표계\\
좌표계 x, y, z와 좌표계 x', y', z'을 보자 두 좌표계의 원점은 같다. 각각의 경우에 대해 벡터 $\mathbf{r}$.은 두 좌표계에서 다음과 같이 표시된다.\\
\\
$\mathbf{r} = x\hat x  + y\hat y  + z\hat z  $. (x, y, z 좌표)\\
\\
$\mathbf{r} = x'\hat x'  + y'\hat y'  + z'\hat z'  $. (x', y', z' 좌표계)\\
\\
벡터의 내적을 이용해 x, y, z를 ($\ x', \hat x', \hat x $.), ($\ y', \hat y', \hat y $.), ($\ z', \hat z', \hat z $.)으로 표현할 수 있다. 내적의 방법은 다음과 같다.\\
\\
$  \mathbf{r}{\hat x} =x = (x'\hat x'  + y'\hat y'  + z'\hat z')(\hat x) = x'(\hat x' \hat x)  + y'(\hat y' \hat x)  + z'(\hat z' \hat x) $.\\
\\
$  \mathbf{r} {\hat y} =y = (x'\hat x'  + y'\hat y'  + z'\hat z')(\hat y) = x'(\hat x' \hat y)  + y'(\hat y' \hat y)  + z'(\hat z' \hat y) $.\\
\\
$  \mathbf{r} {\hat z} =z = (x'\hat x'  + y'\hat y'  + z'\hat z')(\hat z) = x'(\hat x' \hat z)  + y'(\hat y' \hat z)  + z'(\hat z' \hat z) $.\\
\\
으로 표현되는 것을 확인할 수 있다.
\end{CJK}
\end{tcolorbox}
}

\caption{
    Example of suboptimal and our improved extraction from open-source Wikipedia dataset (\url{https://huggingface.co/datasets/wikimedia/wikipedia}).
    The original content is from the Korean Wikipedia article on the Coriolis effect.
}
\end{figure}
\clearpage
\subsection{Details of pre-training from scratch}\label{appendix:pretrain-details}

To control the effects of architecture and tokenization, and to focus on improving the data scaling curve, we adopt the architecture and tokenizer of Llama 3 \citep{llama3}.
Note that while we use the Llama 3 tokenizer, we do not utilize either the weights or the outputs of Llama 3 during the training of Kanana.
Based on the observations of \citet{small-scale-proxy}, we adopt independent weight decay, which follows the original proposal of \citet{adamw} and differs from the PyTorch implementation, and a z-loss \citep{palm} to obtain effective and stable training across various model scales.
We set an independent weight decay of $1\times10^{-4}$ and a z-loss coefficient of $5 \times 10^{-6}$, regardless of model size.
For peak learning rates, learning rate schedulers, and batch sizes, the hyperparameter scaling law and multi-step scheduler from \citet{deepseek-llm} are employed.
\subsection{Details of pruning and distillation}\label{appendix:pd-details}
 
The hyperparameters differ from those used in pre-training from scratch. We apply a cosine learning rate schedule \citep{loshchilov2017sgdr} with an initial learning rate of $1.2 \times 10^{-4}$, batch size of 512, sequence length of 8192, and a warmup phase of 100 steps.
Following the recommendation of Minitron \citep{muralidharan2024compact, sreenivas2024llm}, we employ KL divergence \citep{Kullback1951OnIA} on final logits as the sole loss function.
Additionally, we conclude training early during ablation studies, as pruned models quickly regain performance and the ranking of ablation options rapidly stabilizes.

\begin{table}[h]
\centering
\resizebox{\columnwidth}{!}{
\begin{tabular}{l|cccccc|c}
\toprule
    \multirow{2}{*}{\textbf{Models}} & \textbf{MMLU} & \textbf{KMMLU} & \textbf{HAE-RAE} & \textbf{HumanEval} & \textbf{MBPP} & \textbf{GSM8K} & \multirow{2}{*}{\textbf{Avg}} \\
    & \textit{5-shot} & \textit{5-shot} & \textit{5-shot} & \textit{0-shot} & \textit{3-shot} & \textit{5-shot} & \\
\midrule
Kanana 4.5B &  59.74 &  48.09 &  82.58 &  34.76 &  48.60 &  57.01 &  55.13 \\ 
\midrule
Kanana 3B &  58.21 &  \bfseries 47.55 &  \bfseries79.19 &  34.15 &  45.90 &  53.75 &  53.13 \\
Llama 3.2 3B & 56.40 & 35.57 & 47.66 & 25.61 & 39 & 27.37 & 38.60 \\
Qwen2.5 3B & \bfseries 65.57 & 45.28 & 61.32 & \bfseries 37.80 & \bfseries 55.60 & \bfseries 69.07 & \bfseries 55.77 \\
\midrule
Kanana 2.1B &  54.83 &  \bfseries 44.80 &  \bfseries 77.09 &  31.10 &  \bfseries 46.20 &  46.32 &  \bfseries 50.06 \\
Kanana 1.3B &  53.55 &  39.91 &  72.59 &  28.05 &  39.60 &  36.01 &  44.95 \\
Gemma 2 2B & 52.89 & 30.67 & 45.55 & 20.12 & 28.20 & 24.72 & 33.69 \\
SmolLM2-1.7B & 50.08 & 24.36 & 30.52 & 0.61 & 34.00 & 32.00 & 28.60 \\
Qwen2.5 1.5B & \bfseries 60.86 & 36.63 & 49.68 & \bfseries 37.20 & 44.00 & 62.09 & 48.41 \\
Llama 3.2 1B & 31.51 & 26.46 & 23.10 & 18.90 & 27.60 & 6.14 & 22.29 \\
\midrule
Kanana 635M &  46.28 &  \bfseries 34.60 &  \bfseries 62.69 &  23.17 & \bfseries 31.40 & 19.26 & \bfseries 36.23 \\
Kanana 385M &  41.16 &  31.70 &  47.94 &  18.90 &  24.00 &  10.83 &  29.09 \\
Kanana 192M &  26.11 &  30.16 &  19.71 &  12.80 &  12.40 &  2.43 &  17.27 \\ 
Qwen2.5 0.5B & \bfseries 47.59 & 31.79 & 31.44 & \bfseries 28.66 & 31.00 & \bfseries 35.10 & 34.26 \\ 
SmolLM2-360M & 24.84 & 15.14 & 21.26 & 0.00 & 19.00 & 3.94 & 14.03 \\
SmolLM2-135M & 25.28 & 25.73 & 20.71 & 0.00 & 3.40 & 1.29 & 12.74 \\
\bottomrule
\end{tabular}
}
\caption{
Performance of our models obtained with iterative pruning \& distillation, compared to similar-sized open-source base models.
}\label{tab:pd-base-all}
\end{table}

\begin{figure}[h]
    \centering
    \includegraphics[width=0.7\textwidth]{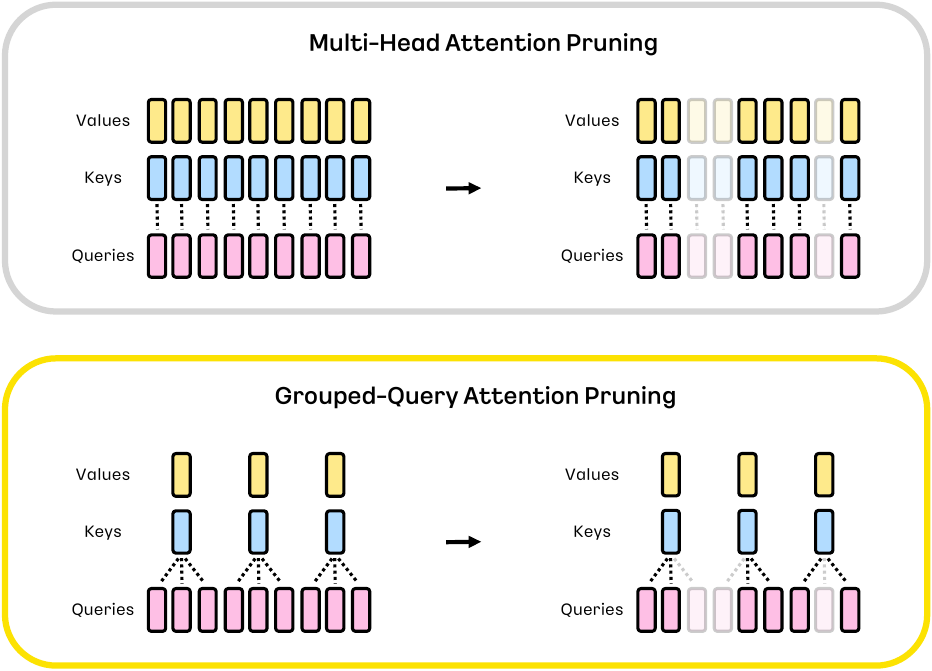}
    \caption{Illustration of ensuring query-key-value alignment in GQA pruning.}
    \label{fig:pd-gqa}
\end{figure}

\begin{table}[h]
    \centering
    \resizebox{0.85\textwidth}{!}{
        \begin{tabular}{cc|ccc|c}
        \toprule
        \multirow{2}{*}{\textbf{GQA alignment}} & \multirow{2}{*}{\textbf{Swiglu importance}} & \multicolumn{3}{c|}{\textbf{Aggregation}} & \multirow{2}{*}{\textbf{Avg}} \\ 
        & & \small{Layer} & \small{Batch} & \small{Sequence} & \\
        \midrule
        \textbf{\cmark} & \textbf{intermediate states} & \textbf{sum} & \textbf{l2norm} & \textbf{avg} & \textbf{36.41} \\
        \textcolor{red}\xmark & intermediate states & sum & l2norm & avg & 20.13 \\
        \cmark & avg of gate, up states & sum & l2norm & avg & 36.04 \\
        \cmark & intermediate states & \textcolor{red}\xmark & l2norm & avg & 13.81 \\
        \cmark & intermediate states & sum & avg & avg & 35.65\\
        \cmark & intermediate states & sum & l2norm & l2norm & 34.25 \\
        \bottomrule
        \end{tabular}
    }
    \caption{Ablation study on importance scoring details, followed by training the same 1.3B architecture with 25B tokens.}
    \label{tab:pd-score-detail}
\end{table}

\begin{table}[h]
    \centering
    \resizebox{0.75\textwidth}{!}{
        \begin{tabular}{cccc|c}
        \toprule
        \multirow{2}{*}{\textbf{Hidden}} & \multirow{2}{*}{\textbf{Intermediate}} & \multirow{2}{*}{\textbf{Query heads}} & \textbf{Non-embedding} & \multirow{2}{*}{\textbf{Avg}} \\
         & & & \textbf{parameters} & \\
        \midrule
        1280 & 5120 & \textbf{24} & 0.96B & \textbf{32.81} \\
        1280 & 5760 & 16 & 0.96B & 28.71 \\
        1280 & 5760 & \textbf{24} & 1.04B & \textbf{34.27}\\
        1536 & 4608 & 16 & 0.98B & 31.99 \\
        1536 & 4608 & \textbf{24} & 1.08B & \textbf{35.10} \\
        1536 & 5376 & 8  & 0.99B & 24.39 \\
        1536 & 5376 & 16 & 1.09B & 32.39 \\
        1536 & 6144 & 8  & 1.11B & 25.91 \\
        \midrule
        1024 & 3072 & 24$\xrightarrow{}$16 & 1.08B$\xrightarrow{}$504M & 20.85 \\
        1024 & 3072 & \textbf{16$\xrightarrow{}$16} & 1.09B$\xrightarrow{}$504M & \textbf{21.78} \\
        \bottomrule
        \end{tabular}
    }
    \caption{Ablation study on model architectures, using 25B training tokens.}
    \label{tab:pd-attn-heads}
\end{table}

\begin{table}[h]
    \centering
    \resizebox{\columnwidth}{!}{
    \begin{tabular}{l|cccccc|c}
    \toprule
    \multirow{2}{*}{\textbf{Embedding}} & \textbf{MMLU} & \textbf{KMMLU} & \textbf{HAERAE} & \textbf{HumanEval} & \textbf{MBPP} & \textbf{GSM8K} & \multirow{2}{*}{\textbf{Avg}}\\
    & \textit{5-shot} & \textit{5-shot} & \textit{5-shot} & \textit{0-shot} & \textit{3-shot} & \textit{5-shot} & \\
\midrule
    tied & 49.07 & 40.41 & 70.49 & 30.49 & 40.60 & 38.21 & 44.88 \\
    untied & 49.88 & 39.61 & 70.21 & 29.88 & 40.20 & 36.92 & 44.45 \\
    \bottomrule
    \end{tabular}
    }
    \caption{Ablation study on tying input and output embeddings by averaging, using 63B training tokens. The rest of the architecture remains unchanged, with 1.86B non-embedding parameters.}
    \label{tab:pd-tie}
\end{table}

\begin{table}[h]
    \centering
    \resizebox{\columnwidth}{!}{
    \begin{tabular}{ll|cccccc|c}
    \toprule
    \multirow{2}{*}{\textbf{Models}} & \multirow{2}{*}{\textbf{Data}}& \textbf{MMLU} & \textbf{KMMLU} & \textbf{HAERAE} & \textbf{HumanEval} & \textbf{MBPP} & \textbf{GSM8K} & \multirow{2}{*}{\textbf{Avg}}\\
    & & \textit{5-shot} & \textit{5-shot} & \textit{5-shot} & \textit{0-shot} & \textit{3-shot} & \textit{5-shot} & \\
\midrule
    1.3B & stage2 & 39.52 & 26.65 & 49.77 & \textbf{25.00} & 32.40 & \textbf{21.00} & 32.39\\
    1.3B & \textbf{stage2 en++} & \textbf{44.00} & \textbf{33.90} & \textbf{62.42} & 23.78 & \textbf{33.80} & 20.55 & 
    \textbf{36.41}\\
    \bottomrule
    \end{tabular}
    }
    \caption{Ablation study on distillation data, using 25B training tokens.}
    \label{tab:pd-distill-data}
\end{table}

\clearpage

\section{Evaluation Details}

\subsection{Evaluation prompts for post-trained models} \label{subsec:evaluation-prompts-for-post-trained-models}

\begin{figure}
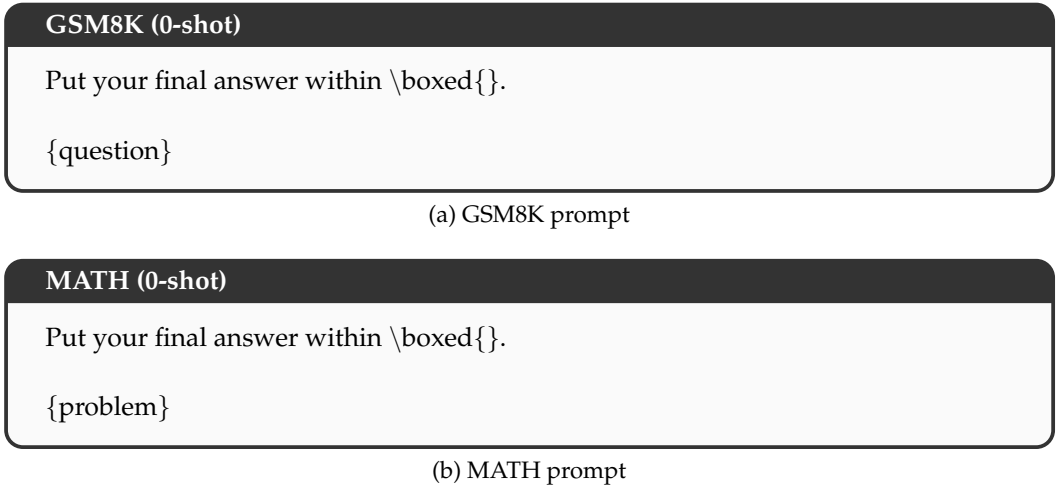

    \centering
    \subfloat[GSM8K prompt]{
    \begin{tcolorbox}[colframe=black!80!white, colback=black!2!white, boxrule=0.5mm, width=\textwidth, arc=2mm, auto outer arc, title=GSM8K (0-shot), fonttitle=\color{white}\bfseries]
    \setstretch{1.2}
Put your final answer within \textbackslash boxed\{\}.\\\\
\{question\}
    \end{tcolorbox}
        \label{fig:gsm8k-zero-shot-prompt}
    }
    
    \subfloat[MATH prompt]{
    \begin{tcolorbox}[colframe=black!80!white, colback=black!2!white, boxrule=0.5mm, width=\textwidth, arc=2mm, auto outer arc, title=MATH (0-shot), fonttitle=\color{white}\bfseries]
    \setstretch{1.2}
Put your final answer within \textbackslash boxed\{\}.\\\\
\{problem\}
    \end{tcolorbox}
        \label{fig:math-zero-shot-prompt}
    }
    
    \caption{
    Evaluation prompts for GSM8K and MATH.
    The prompts are used to evaluate instruction-tuned models.
    }
    \label{fig:evaluation-prompts-math}
\end{figure}

We employ 0-shot CoT prompts for a number of evaluations. 
See \autoref{fig:evaluation-prompts-general} for MMLU, KMMLU and HAE-RAE.
For math-related tasks we employ 0-shot prompt.
We refer the readers to \autoref{fig:evaluation-prompts-math} for the prompts corresponding to GSM8K and MATH.

\begin{figure}
    \centering
    \subfloat[MMLU prompt]{
        \begin{tcolorbox}[colframe=black!80!white, colback=black!2!white, boxrule=0.5mm, width=\textwidth, arc=2mm, auto outer arc, title=MMLU prompt (0-shot CoT), fonttitle=\color{white}\bfseries]
            \setstretch{1.2}
            The following are multiple choice questions about \texttt{\{mmlu\_subject\}}. Summarize your reasoning concisely, then conclude with "Therefore, the answer is: X" where X is one of A, B, C, or D.\\
            
            Question: \texttt{\{question\}}\\
            A. \texttt{\{choice\_A\}}\\
            B. \texttt{\{choice\_B\}}\\
            C. \texttt{\{choice\_C\}}\\
            D. \texttt{\{choice\_D\}}\\
        \end{tcolorbox}
        \label{fig:mmlu-zero-shot-cot-prompt}
    }
    
    \subfloat[KMMLU prompt]{
        \begin{tcolorbox}[colframe=black!80!white, colback=black!2!white, boxrule=0.5mm, width=\textwidth, arc=2mm, auto outer arc, title=KMMLU prompt (0-shot CoT), fonttitle=\color{white}\bfseries]
            \setstretch{1.2}
            \begin{CJK}{UTF8}{mj}
            다음은 \texttt{\{kmmlu\_subject\}}에 관한 객관식 문제입니다. 당신의 추론 과정을 간결하게 요약한 후, "따라서, 정답은: X"라고 결론지으십시오. 여기서 X는 A, B, C, D 중 하나입니다.\\
            
            질문: \texttt{\{question\}}\\
            A. \texttt{\{choice\_A\}}\\
            B. \texttt{\{choice\_B\}}\\
            C. \texttt{\{choice\_C\}}\\
            D. \texttt{\{choice\_D\}}\\
            \end{CJK}
        \end{tcolorbox}
        \label{fig:kmmlu-zero-shot-cot-prompt}
    }
    
    \subfloat[HAE-RAE prompt]{
        \begin{tcolorbox}[colframe=black!80!white, colback=black!2!white, boxrule=0.5mm, width=\textwidth, arc=2mm, auto outer arc, title=HAE-RAE (0-shot CoT), fonttitle=\color{white}\bfseries]
        \setstretch{1.2}
    \begin{CJK}{UTF8}{mj}
    다음은 객관식 문제입니다. 당신의 추론 과정을 간결하게 요약한 후, "따라서, 정답은: X"라고 결론지으십시오. 여기서 X는 A, B, C, D, E 중 하나입니다.\\
    \{query\}
    \end{CJK}
        \end{tcolorbox}
        \label{fig:haerae-zero-shot-cot-prompt}
    }
    
    \caption{
    Evaluation prompts for MMLU, KMMLU, and HAE-RAE.
    The prompts are used to evaluate instruction-tuned models.
    }
    \label{fig:evaluation-prompts-general}
\end{figure}

\section{Qualititive Results} \label{sec:qualititive_results}

In this section, we include qualitative experimental results.
We mainly focus on changes made during post-training.
See \autoref{tab:completions_during_post_train_1} and \autoref{tab:completions_during_post_train_2} for representative examples.

\begin{CJK}{UTF8}{mj}

\begin{longtable}{lp{0.8\textwidth}}
\toprule
\textbf{Prompt} & \textbf{1 더하기 1은 귀요미야! 1 더하기 1은?} \\
\midrule
\endfirsthead

\multicolumn{2}{c}{\tablename\ \thetable\ -- \textit{Continued from previous page}} \\
\toprule
\textbf{Prompt} & \textbf{1 더하기 1은 귀요미야! 1 더하기 1은?} \\
\midrule
\endhead

\midrule
\multicolumn{2}{r}{\textit{Continued on next page}} \\
\endfoot

\bottomrule
\caption{Prompt and completions of each post-training stage.}
\label{tab:completions_during_post_train_1} \\
\endlastfoot

SFT & 1 더하기 1은 2입니다. \\\\

DPO & 1 더하기 1은 2입니다. 귀요미라는 말은 상황에 따라 다르게 해석될 수 있지만, 일반적으로는 '귀여운 것'을 의미합니다. 예를 들어, 아이들이 수학 문제를 재미있게 풀기 위해 "1 더하기 1은 귀요미" 같은 구호를 사용하기도 합니다. 하지만 수학적으로는 1 + 1 = 2입니다. \\\\
Online DPO & 1 더하기 1은 2입니다. 귀요미라는 표현은 수학적인 계산보다는 감정적인 표현으로 보입니다. 혹시 다른 질문이나 추가적인 설명이 필요하시면 알려주세요! \\

\end{longtable}

\begin{longtable}{lp{0.8\textwidth}}
\toprule
\textbf{Prompt} & \textbf{
Convert these dates to YYYY/MM/DD format:\newline
12/31/2021\newline
31-12-21\newline
01/02/2022\newline
02-01-22\newline
} \\
\midrule
\endfirsthead

\multicolumn{2}{c}{\tablename\ \thetable\ -- \textit{Continued from previous page}} \\
\toprule
\textbf{Prompt} & \textbf{
Convert these dates to YYYY/MM/DD format:\newline
12/31/2021\newline
31-12-21\newline
01/02/2022\newline
02-01-22\newline
} \\
\midrule
\endhead

\midrule
\multicolumn{2}{r}{\textit{Continued on next page}} \\
\endfoot

\bottomrule
\caption{Prompt and completions of each post-training stage.}
\label{tab:completions_during_post_train_2} \\
\endlastfoot

SFT & Here is a Python solution for the problem:

\begin{lstlisting}[language=Python, basicstyle=\scriptsize]
def convert_date(date):
    if '-' in date:
        date_parts = date.split('-')
        return f"{date_parts[2]}/{date_parts[1]}/{date_parts[0]}"
    else:
        date_parts = date.split('/')
        return f"{date_parts[2]}/{date_parts[0]}/{date_parts[1]}"
 
dates = ["12/31/2021", "31-12-21", "01/02/2022", "02-01-22"]
for date in dates:
    print(convert_date(date))
\end{lstlisting}

When you run this code, it will print: \newline\newline
2021/12/31\newline
2021/12/31\newline
2022/02/01\newline
2022/01/02\newline
This code works by checking if the input date contains a '-' character.
If it does, it assumes the date is in the format DD-MM-YYYY and converts it to YYYY/MM/DD.
If the input date does not contain a '-' character, it assumes the date is in the format MM/DD/YYYY and converts it to YYYY/MM/DD.

\\\\

DPO & Here is a Python solution for the problem:

\begin{lstlisting}[language=Python, basicstyle=\scriptsize]
def convert_date(date):
    if '/' in date:
        month, day, year = date.split('/')
        return f"{year}/{month.zfill(2)}/{day.zfill(2)}"
    elif '-' in date:
        day, month, year = date.split('-')
        return f"{year}/{month.zfill(2)}/{day.zfill(2)}"
    else:
        raise ValueError("Invalid date format")
 
dates = ["12/31/2021", "31-12-21", "01/02/2022", "02-01-22"]
 
for date in dates:
    print(convert_date(date))
\end{lstlisting}
This script will convert the dates to the desired format. The zfill(2) function is used to add leading zeros if the month or day is a single digit.\newline\newline
 
When you run this script, it will output:\newline
\newline
\newline
2021/12/31\newline
2021/12/31\newline
2022/01/02\newline
2022/01/02\newline
Note that the dates "31-12-21" and "02-01-22" are interpreted as "31 December 2021" and "2 January 2022" respectively, because the year is given as "21" and "22". If these dates are meant to represent "12-31-2021" and "1-2-2022" (i.e., December 31, 2021 and January 2, 2022), then the script will need to be modified to handle this interpretation.
\\\\
Online DPO & 
Sure, here are the dates converted to the YYYY/MM/DD format:\newline\newline
12/31/2021 $\rightarrow$ 2021/12/31\newline
31-12-21 $\rightarrow$ 2021/12/31\newline
01/02/2022 $\rightarrow$ 2022/01/02\newline
02-01-22 $\rightarrow$ 2022/01/02\newline
Note that the dates "31-12-21" and "02-01-22" are ambiguous and could potentially mean the opposite year, but based on common sense and the context provided, the most likely interpretation is the one given above (i.e., 2021 and 2022 respectively). If the exact year is unknown, it's crucial to have additional context.
\\

\end{longtable}

\end{CJK}

\section{Evaluation Details of Embedding Models}\label{appendix:embedding}
\begin{table}[H]
    \centering
    \resizebox{\columnwidth}{!}{%
    \begin{tabular}{l|ccccc}
        \toprule
        \textbf{Benchmarks} & \textbf{Kanana-Nano-2.1B} & \textbf{Llama3.2 3B} & \textbf{Qwen2.5 3B} & \textbf{Llama3.2 1B} & \textbf{Qwen2.5 1.5B} \\
        \midrule
        \midrule
        \textbf{English (Avg)} & 51.56 & 53.28 & 54.00 & 48.77 & 50.60 \\
        \midrule
        ArguAna & 54.59 & 54.36 & 56.26 & 51.80 & 53.49 \\
        CQADupstackGamingRetrieval & 58.37 & 60.31 & 59.65 & 56.13 & 57.20 \\
        CQADupstackUnixRetrieval & 43.34 & 45.27 & 45.41 & 39.18 & 41.14 \\
        ClimateFEVERHardNegatives & 29.64 & 30.64 & 31.08 & 26.93 & 27.66 \\
        FEVERHardNegatives & 73.18 & 79.09 & 80.26 & 73.27 & 72.09 \\
        FiQA2018 & 40.22 & 46.47 & 47.12 & 38.54 & 41.08 \\
        HotpotQAHardNegatives & 61.35 & 66.10 & 66.33 & 61.21 & 64.18 \\
        SCIDOCS & 21.41 & 21.44 & 22.14 & 18.96 & 19.81 \\
        TRECCOVID & 79.85 & 81.84 & 80.87 & 72.67 & 75.88 \\
        Touche2020Retrieval.v3 & 53.63 & 47.26 & 50.91 & 49.00 & 53.50 \\
        \midrule
        \midrule
        \textbf{Korean (Avg)} & 65.00 & 59.43 & 62.10 & 54.68 & 54.60 \\
        \midrule
        AutoRAGRetrieval & 79.71 & 70.87 & 75.64 & 71.47 & 72.32 \\
        BelebeleRetrieval & 92.35 & 87.58 & 90.16 & 84.44 & 83.53 \\
        Ko-StrategyQA & 79.98 & 73.92 & 76.38 & 63.46 & 64.97 \\
        MIRACLRetrieval & 60.04 & 52.25 & 56.83 & 48.28 & 48.68 \\
        MrTidyRetrieval & 49.82 & 45.83 & 48.48 & 35.32 & 37.94 \\
        MultiLongDocRetrieval & 30.17 & 25.54 & 25.75 & 20.98 & 17.13 \\
        PublicHealthQA & 88.08 & 84.12 & 86.68 & 80.26 & 79.71 \\
        XPQARetrieval & 39.88 & 35.33 & 36.89 & 33.24 & 32.55 \\
        \bottomrule
    \end{tabular}%
    }
    \caption{Evaluation details of embedding models on English and Korean retrieval benchmarks.}
    \label{tab:embedding-details}
\end{table}
\section{RAG-General-Bench Examples}\label{appendix:rag-bench-example}

\begin{CJK}{UTF8}{mj}

\begin{figure}[h]
    \centering
    \subfloat[][Sample 1]{%
        \fbox{\includegraphics[width=0.45\textwidth]{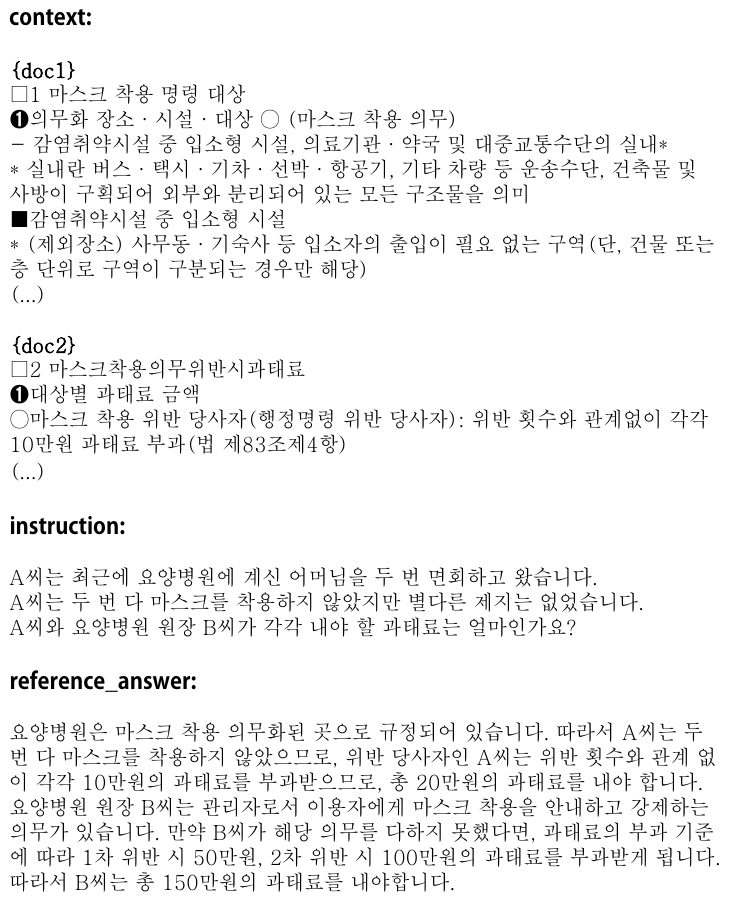}}%
        \label{fig:rag_pre_stage1_data}
    }
    \hfill
    \subfloat[][Sample 2]{%
        \fbox{\includegraphics[width=0.45\textwidth]{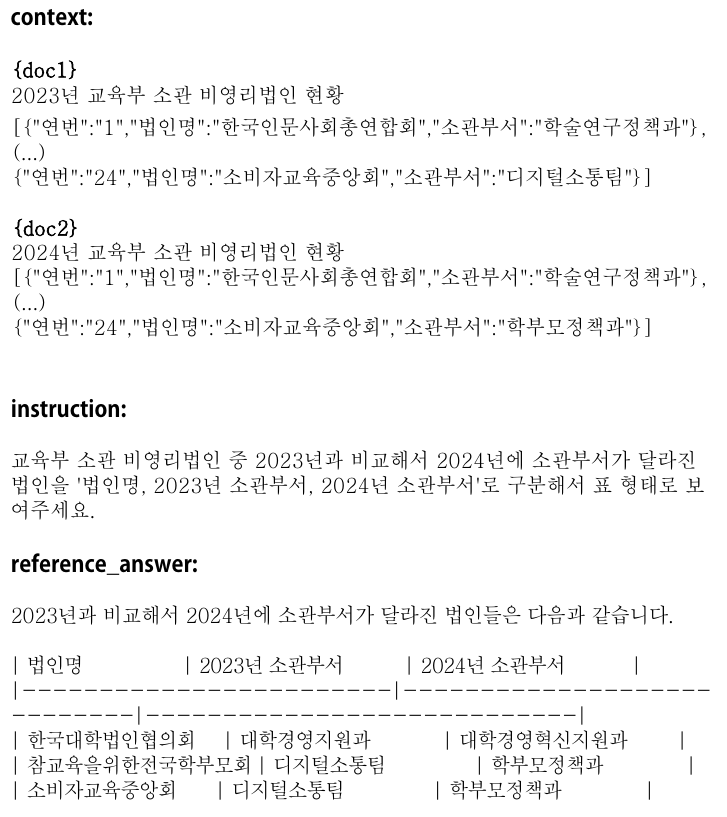}}%
        \label{fig:rag_pre_stage2_data}
    }
    \caption{RAG-General-Bench example: QA}
    \label{fig:rag_sample1}
\end{figure}

\end{CJK}

\section{FunctionChat-Bench Examples}\label{appendix:fc-bench-example}
\subsection{Single-call}
\begin{figure}[h]
    \centering
    \resizebox{0.7\textwidth}{!}{%
        \includegraphics[width=\textwidth]{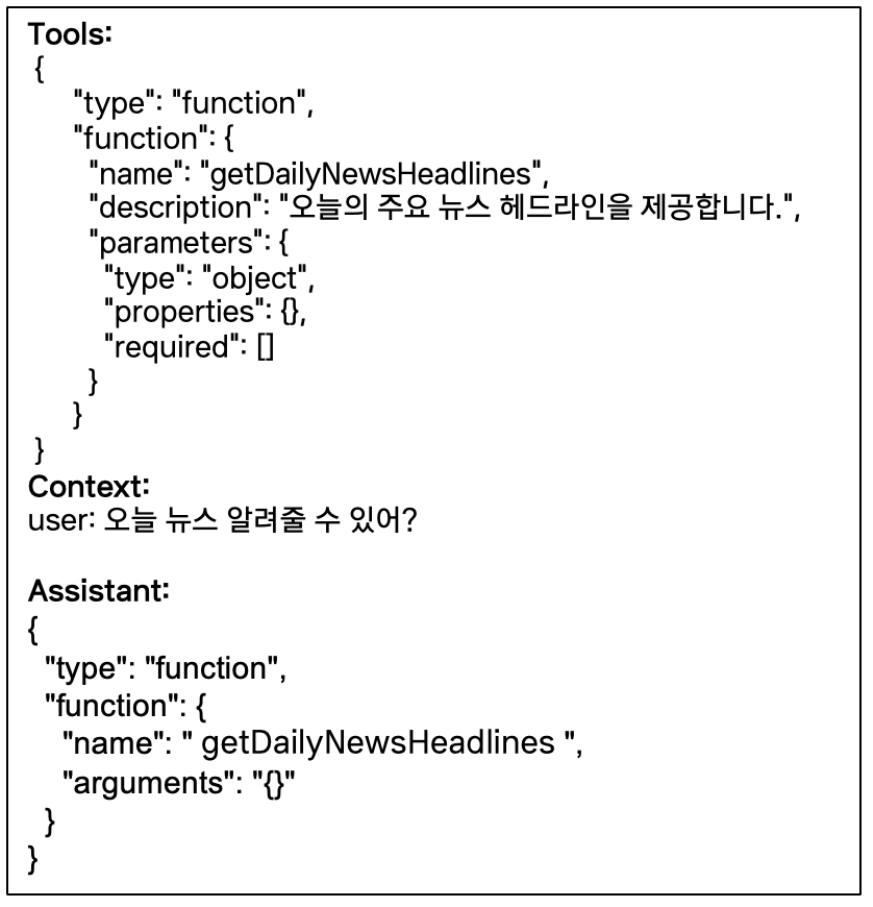}
    }
    \caption{FunctionChat-bench example : Single-call(1\_exact)}
    \label{fig:fc-appendix-single}
    \end{figure}

Single-call evaluates how accurately the LM can select and call the necessary function among several options by providing four single-turn prompts for each of 25 different functions. As show in Figure \ref{fig:fc-appendix-single}, "1\_exact" is that only the target function is provided to the Assistant as a candidate.

\subsection{Dialogue}
The dialog dataset consists of 45 diverse multi-turn interactions between real users and an LM, categorized into four situation types to evaluate the model's response accuracy and appropriateness.
\begin{enumerate}
    \item \textbf{Call}: An LM must accurately select functions and extract the necessary parameters to respond to a user prompt
    \item \textbf{Completion}: An LM must generate appropriate responses based on the results of the tool.
    \item \textbf{Slot}: An LM must query the user for the necessary parameters to make a function call.
    \item \textbf{Relevance}: An LM must generate an appropriate response when it cannot provide a function for a user prompt.
\end{enumerate}

    \begin{figure}[h]
    \centering
    \resizebox{1\textwidth}{!}{%
        \includegraphics[width=\textwidth]{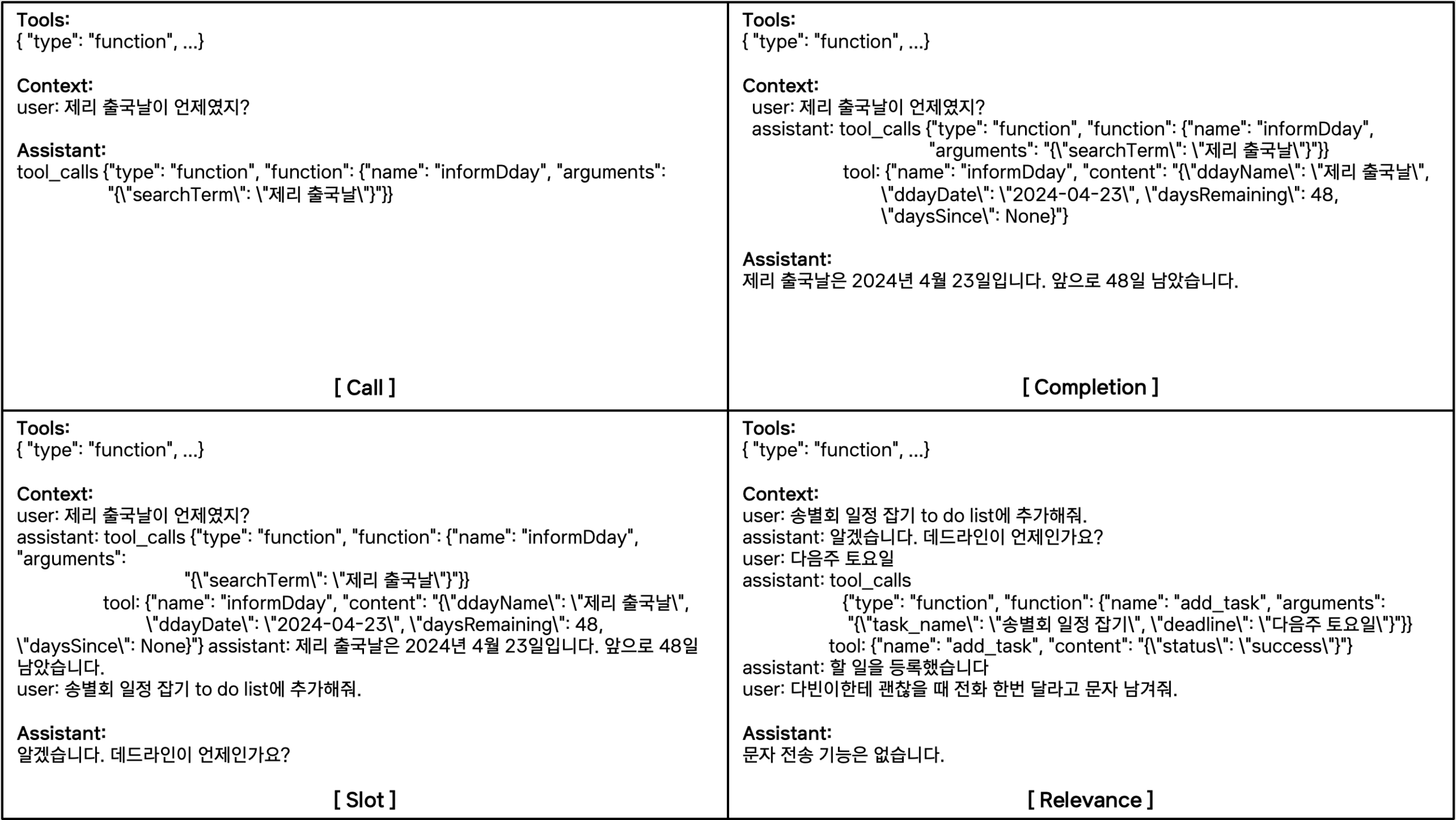}
    }
    \caption{FunctionChat-bench Example : Dialogue}
    \label{fig:fc-appendix-dialogue}
    \end{figure}

\end{document}